%% file: main.tex
\documentclass{article}

\usepackage{microtype}
\usepackage{graphicx}
\usepackage{subcaption}
\usepackage{booktabs}
\usepackage{xurl}
\usepackage{hyperref}

\let\origaddcontentsline\addcontentsline
\usepackage[preprint]{icml2026}
\usepackage{amsmath}
\usepackage{amssymb}
\usepackage{mathtools}
\usepackage{amsthm}
\usepackage[capitalize,noabbrev]{cleveref}

\input{sections/tables/macros}

\usepackage{etoc}

\DeclareMathOperator{\TV}{TV}
\DeclareMathOperator{\KL}{KL}
\DeclareMathOperator{\EAL}{EAL}
\DeclareMathOperator{\E}{E}

\theoremstyle{plain}
\newtheorem{theorem}{Theorem}[section]
\newtheorem{proposition}[theorem]{Proposition}
\newtheorem{lemma}[theorem]{Lemma}

\theoremstyle{definition}

\theoremstyle{remark}

\AddToHook{env/proposition/begin}{\crefalias{theorem}{proposition}}
\AddToHook{env/lemma/begin}{\crefalias{theorem}{lemma}}
\AddToHook{env/corollary/begin}{\crefalias{theorem}{corollary}}
\AddToHook{env/definition/begin}{\crefalias{theorem}{definition}}
\AddToHook{env/assumption/begin}{\crefalias{theorem}{assumption}}
\AddToHook{env/remark/begin}{\crefalias{theorem}{remark}}
\crefname{assumption}{Assumption}{Assumptions}
\Crefname{assumption}{Assumption}{Assumptions}

\icmltitlerunning{TVP: TV is Almost All You Need for Speculative Decoding}
\icmlkeywords{speculative decoding, total variation, language models}

\begin{document}

\twocolumn[
  \icmltitle{TVP: TV is Almost All You Need for Speculative Decoding}
  \begin{icmlauthorlist}
    \icmlauthor{Aaryam Sharma}{baseten}
  \end{icmlauthorlist}
  \icmlaffiliation{baseten}{Baseten Labs Inc., \url{https://www.baseten.co/}}
  \icmlcorrespondingauthor{Aaryam Sharma}{aaryam.sharma@baseten.co}
  \vskip 0.3in
]

\printAffiliationsAndNotice{}

\begin{abstract}

  Speculative decoding accelerates LLM inference with a drafter whose proposed tokens are verified by the target model in parallel. Under lossless speculative sampling, a draft token is accepted with probability one minus the total variation (TV) distance between the target and draft distributions, yet drafters are often trained with cross-entropy (CE) or KL. We first show that under matched and well-tuned training, TV matches or surpasses several published draft-training objectives when fine-tuning, and a fixed TV and KL mixture does the same when training from scratch. We identify two remaining problems, TV's vanishing gradients on undercovered tokens and the prefix products that starve later positions in acceptance-length objectives. To address both, we derive TVP, a loss family based on expected acceptance length, whose power parameter interpolates between coverage and acceptance oriented objectives. Annealing the power from zero to one gives the highest or tied-highest acceptance length across three drafter architectures and three target models, including longer drafts and block verification. TVP improves acceptance length over CE by 9.8\% when fine-tuning, and over the public DFlash2 training recipes by 6.2 to 12.9\%.

\end{abstract}

\etocdepthtag.toc{mtmain}
\input{sections/introduction}
\input{sections/related_work}
\input{sections/background}

\input{sections/method}
\input{sections/theoretical_properties}
\input{sections/experiments}

\input{sections/conclusion}
\input{sections/acknowledgements}
\input{sections/impact_statement}

\input{sections/references}

\clearpage
\appendix
\onecolumn
\let\addcontentsline\origaddcontentsline
\etocdepthtag.toc{mtappendix}
\etocsettagdepth{mtmain}{none}
\etocsettagdepth{mtappendix}{subsection}
\etocsettocstyle{\section*{Appendix Contents}}{}
\tableofcontents
\clearpage
\input{sections/appendix/appendix}

\end{document}

%% file: sections/tables/macros.tex
\newcommand{\ci}[1]{\,{\scriptsize$\pm$#1}}
\newcommand{\spd}[2]{#1$\times$\ci{#2}}
\newcommand{\tvref}{\textit{ref}}

\newcommand{\anneal}[1]{$\beta$\,#1$\to$1}

%% file: sections/introduction.tex
\section{Introduction}
\label{sec:introduction}

Speculative decoding addresses the bottleneck of autoregressive generation in LLMs by introducing a light draft model that proposes tokens, which the larger target model verifies in parallel. Thus, instead of generating tokens one at a time, the target model can output multiple tokens at once. With an appropriate verification mechanism, speculative decoding preserves the target model's original output distribution~\citep{leviathan23a,chen2023accelerating}. The drafter can be a smaller language model or a lightweight head conditioned on the target model's hidden states. Its effectiveness depends on the cost of drafting, target verification, and number of accepted tokens.

Draft training often uses cross-entropy or divergence-based distillation \citep{DistillSpec}. Under speculative sampling, however, the acceptance probability is $1-\TV(p,q)$, where $p$ and $q$ are the target and draft distributions and TV denotes their total variation distance. Minimizing TV hence directly relates to maximizing the probability of acceptance at each prefix~\citep{leviathan23a,SpecDecTheory}.

TV has two limitations. Within each position, it assigns little corrective weight to severely undercovered target tokens, making it difficult to correct a draft that is initially poorly aligned with the target~\citep{LKLoss}. Across positions, an additive TV objective ignores that acceptance at each position depends on the preceding prefix surviving verification. For a fixed draft architecture and length, improving expected acceptance length (EAL) means generating more tokens for approximately the same drafting and verification cost. Under realizability, minimizing TV at every position also maximizes EAL, as both are optimized by $q=p$. A drafter with limited capacity, however, must trade off errors across positions. This motivates optimizing EAL itself, but its gradients contain prefix products that weaken supervision at later positions when earlier tokens are rarely accepted.

We introduce TVP, a new family of losses derived from EAL that addresses both limitations. TVP replaces prefix products with additive supervision across draft positions and introduces a power parameter that controls the corrective weight assigned to undercovered tokens. Starting with stronger undercoverage correction helps train an initially misaligned draft. Annealing the power parameter toward one then shifts the objective toward acceptance as training progresses, keeping TV's direct acceptance signal once the draft is aligned. We establish bounds connecting this family to EAL and local rejection probability, and show that annealing recovers both TV's loss and its gradients.

In our main experiments, we compare TVP with a wide range of existing objectives using DSpark, both when fine-tuning and when training from scratch, while also exploring Eagle3 and DFlash2. We additionally examine the learning rate, training budget, target model, and draft length.

We make four contributions.
\begin{itemize}
  \item Through a controlled comparison, we show that under well-tuned training, TV often matches or surpasses more complex objectives when fine-tuning drafters across architectures, while from scratch a small KL component substantially improves EAL.
  \item We extend the gradient analysis of acceptance objectives to any draft and across positions. TV's gradient vanishes whenever the draft puts little mass on undercovered tokens, and EAL's prefix products starve later positions, so optimizing EAL directly is 9.5\% worse than TV from scratch.
  \item We derive TVP from EAL through a composition rule. Its power interpolates between coverage and acceptance, and its value lower-bounds EAL.
  \item We show that annealing the power from zero to one improves on TV by 1.5\% when fine-tuning, on the best published objective by 1.2\% from scratch, on CE by 9.8\%, and on the public DFlash2 recipes by 6.2 to 12.9\%. Across drafters, targets, draft lengths, and block verification~\citep{sun2025block}, TVP gives the highest or tied-highest acceptance length.
\end{itemize}

%% file: sections/related_work.tex
\section{Related Work}
\label{sec:related-work}

\paragraph{Draft architectures.}
\citet{blockwise_parallel_decoding} introduced blockwise parallel decoding for greedy generation. Lookahead decoding constructs drafts without training a draft model~\citep{LookaheadDecoding}. Early speculative sampling methods used a smaller language model as the drafter~\citep{leviathan23a,chen2023accelerating}. Later work introduced draft heads that were target-specific, using representations from the target model directly. Medusa and Hydra use multiple decoding heads, one per position~\citep{ankner2024hydra,Medusa}, while the EAGLE family uses a single small autoregressive head repeatedly~\citep{Eagle1,Eagle2,Eagle3}.

More recent architectures generate drafts in parallel. DFlash uses block-diffusion conditioned on target features~\citep{DFlash}, and PARD-2 generates parallel drafts~\citep{PARD2}. DSpark combines a parallel backbone, based on DFlash, with a small sequential module (Markov head) to model dependencies within the block~\citep{DSpark}. DFlash2 adds dynamic convolutions and a path selector over candidate tokens to model sequential dependencies~\citep{inco2026dflash2}. These architectures provide different settings in which to study a training objective, and common serving frameworks support them efficiently and at scale~\citep{zheng2024sglang,kwon2023vllm}.

\paragraph{Training objectives.}
Draft training is often formulated as distillation from the target model, using cross-entropy or forward KL divergence~\citep{Eagle3,lightseek2026torchspec}. \citet{DistillSpec} studied how the training distribution and the choice of divergence affect speculative decoding. KL-based acceptance guarantees have been studied for greedy, relaxed, and tree-based verification~\citep{GreedyTheory}. AdaSPEC uses token filtering in the distillation process to concentrate the draft model's limited capacity~\citep{hu2025adaspec}, while Draft-OPD applies on-policy distillation to draft training~\citep{DraftOPD}.

TaiLr~\citep{ji2023tailoring} trains language generation models under TV, with a parameter that interpolates between MLE and TV to balance the bias and variance of estimating TV from data samples. \citet{goel2024direct} use a TV-based loss to align draft models. A large number of recent methods incorporate acceptance through adaptive mixtures, position weighting, and sequence-level surrogates. LK losses mix KL and TV with a weight that decreases with the current acceptance at each position, shifting from KL to TV as the draft aligns~\citep{LKLoss}. TVP instead derives a single loss family from EAL, applies it to the full target distribution, and anneals it from coverage to acceptance.

Token acceptance also depends on whether verification reaches that position. Some methods reweight the loss by position. PARD-2's confidence-adaptive token (CAT) weighting uses the cumulative product of target confidence on the token's prefix~\citep{PARD2}. Other methods construct surrogates for acceptance length. D-PACE and its KL variant, D-PAKL, derive position weights from draft predictions~\citep{DPACE}. DBLast uses prefix likelihood ratios in an acceptance-length surrogate alongside its dependent drafting architecture~\citep{DBLAST}. Window TV (WTV) couples token overlaps through prefix products~\citep{WTV}. BV derives its loss from block verification~\citep{sun2025block,BV}. We train DBLast's loss without its latent-variable architecture (DBLast-AL) and apply CAT weighting to TV+KL. \Cref{sec:appendix-losses} defines every objective we compare. Outside speculative decoding, the generalized cross-entropy for noisy labels applies the power form of $\phi_\beta$ to the label probability~\citep{generalized_ce}.

This work examines both the objective and how it is scheduled during training. We distinguish fine-tuning an existing draft head from training one from scratch, and study how the loss schedule affects acceptance in each regime.

%% file: sections/background.tex
\section{Background and Motivation}
\label{sec:background}
\label{sec:motivation}

\subsection{Speculative Sampling}
\label{sec:speculative-sampling}

Let the current context of tokens be $x=x_{1:n}$, and let $p(\cdot\mid x)$ and $q_\theta(\cdot\mid x)$ denote the target and draft distributions over the vocabulary $\mathcal{V}$, respectively. For a fixed context, we omit the context conditional and write $p$ and $q$. For a continuation $Y_{1:K}$ of length $K$, we write $p_i=p(\cdot\mid x,Y_{<i})$ and $q_i=q_\theta(\cdot\mid x,Y_{<i})$.

In this paper, we focus on the target distribution preserving formulation of \citet{leviathan23a}. Draft tokens are sampled from $q$ and verified tokenwise. If all previous draft tokens have been accepted, $y_i$ is accepted with probability
\begin{equation}
  \alpha_i(y_i)=\min\left\{1,\frac{p_i(y_i)}{q_i(y_i)}\right\}.
  \label{eq:speculative-acceptance-rule}
\end{equation}
At the first rejection, subsequent draft tokens are discarded, and a replacement token is sampled from the distribution
\begin{equation}
  p_i^{\mathrm{res}}(v)=\frac{\max\{0,p_i(v)-q_i(v)\}}
  {\sum_{u\in\mathcal{V}}\max\{0,p_i(u)-q_i(u)\}}.
  \label{eq:residual-distribution}
\end{equation}
If all $K$ draft tokens are accepted, an extra token is sampled from $p(\cdot\mid x,y_{1:K})$. This method preserves the target model's output distribution. Alternatively, in greedy decoding, a draft token is accepted if its prefix is accepted and it matches the target's argmax.

The total variation distance (TV) between $p$ and $q$ is
\begin{equation}
  \operatorname{TV}(p,q)
  =\frac{1}{2}\sum_{v\in\mathcal{V}}|p(v)-q(v)|.
  \label{eq:total-variation}
\end{equation}
A standard result shows that under speculative sampling, the per-token acceptance probability is $1-\operatorname{TV}(p,q)$~\citep{leviathan23a,SpecDecTheory}. Specifically,
\begin{equation}
\begin{aligned}
  \alpha(p,q)
  &=\mathbb{E}_{y\sim q}\!\left[\min\left\{1,\frac{p(y)}{q(y)}\right\}\right]\\
  &=\sum_{v\in\mathcal{V}}\min\{p(v),q(v)\}=1-\operatorname{TV}(p,q).
\end{aligned}
\label{eq:acceptance-tv}
\end{equation}
Thus, for a fixed context, minimizing TV is equivalent to maximizing the probability of accepting the next draft token.

\subsection{Expected Acceptance Length}
\label{sec:expected-acceptance-length}

Let $N\le K$ be the number of accepted draft tokens before the first rejection, excluding the correction token. We write $\EAL=\E[N]$ for expected acceptance length. Including the correction token gives the mean number of tokens per verification step, $\tau=1+\EAL$. On a draft-sampled continuation $Y\sim q$, define $A_t=\prod_{i=1}^t\alpha_i(Y_i)$. On a target-sampled continuation $Y\sim p$, define $r_i=\min\{1,q_i(Y_i)/p_i(Y_i)\}$ and $B_t=\prod_{i=1}^t r_i$, with $A_0=B_0=1$. We use $\E_q$ and $\E_p$ for expectations over draft and target continuations.

With limited drafter capacity, minimizing TV or KL locally may not optimize EAL, because local improvements may sacrifice acceptance for later tokens. EAL can be written with prefix products under either the draft or the target,
\begin{equation}
\label{eq:tvp-eal}
\EAL=\sum_{t=1}^{K}\E_q[A_t]=\sum_{t=1}^{K}\E_p[B_t].
\end{equation}
Appendix~\ref{app:tvp-eal} proves this change of measure.

\input{sections/motivation}

%% file: sections/motivation.tex
\subsection{Gradients of TV and KL}
\label{sec:tv-kl-gradients}

\begin{figure}[t]
  \centering
  \includegraphics[width=\columnwidth]{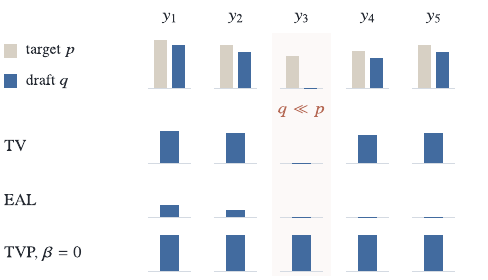}
  \caption{Target and draft probabilities (top) and score-gradient weights divided by $w_i$ (bottom) on one target-sampled continuation. TV barely corrects the undercovered $y_3$, and EAL also attenuates later positions, while TVP at $\beta=0$ retains correction throughout the window.}
  \label{fig:gradients}
\end{figure}

We first compare how TV and KL correct undercovered tokens. We write $U=\{v\mid q(v)<p(v)\}$ for the undercovered tokens and $a=p(U)$, $b=q(U)$ for their target and draft mass. Then $\TV(p,q)=a-b$. We use $\KL(p\|q)=\sum_{v\in\mathcal{V}}p(v)\log(p(v)/q(v))$. For gradient calculations, the target is fixed and $q=q_\theta=\operatorname{softmax}(z)$, where $z$ denotes the draft logits.

The parameter gradients of TV and KL are
\begin{equation}
\label{eq:tvp-tv-kl-grad}
\begin{aligned}
\nabla_\theta\TV(p,q)
  &=-\sum_{v\in U}q(v)\nabla_\theta\log q(v),\\
\nabla_\theta\KL(p\|q)
  &=-\sum_v p(v)\nabla_\theta\log q(v).
\end{aligned}
\end{equation}
Their logit gradients satisfy
\begin{equation}
\label{eq:tvp-tv-kl-norm}
\|\nabla_z\TV(p,q)\|_1=2b(1-b),\enspace
\|\nabla_z\KL(p\|q)\|_1=2(a-b).
\end{equation}
When the draft puts almost no mass on the undercovered tokens ($b\to0$) while the target does ($a>0$), TV's gradient vanishes (\cref{fig:gradients}), while KL's gradient does not. \citet{LKLoss} observe that TV's logit gradient reflects only the sign of each token's error and vanishes at random initialization. \Cref{eq:tvp-tv-kl-norm} makes the condition exact for any draft, including a trained one that is confident in the wrong token. Appendix~\ref{app:tvp-gradients} gives the derivation.

\subsection{Prefix Attenuation}
\label{sec:prefix-attenuation}

Since $p$ does not depend on $\theta$, we can differentiate \cref{eq:tvp-eal} inside $\E_p$. The factor $r_i$ has a nonzero gradient only when $Y_i$ lies in the undercovered set $U_i$ of $q_i$, which gives
\begin{equation}
\label{eq:tvp-attenuation}
\nabla_\theta\frac{\EAL}{K}=\E_p\sum_{i:\,Y_i\in U_i}\Bigl(\frac1K\sum_{t=i}^{K}B_t\Bigr)\nabla_\theta\log q_i(Y_i).
\end{equation}
The coefficient at position $i$ is at most $w_iB_i$, where $w_i=(K-i+1)/K$ (\cref{app:tvp-eal}).
While $w_i$ decays only linearly, $B_i$ can decay exponentially due to multiplicative accumulation of mismatch, diminishing the learning signal for all future tokens (\cref{fig:gradients}). When a head is trained from scratch, the draft probabilities start off as poorly aligned, and the initial learning signal then mostly applies to the first few tokens, slowing down training. Optimizing EAL directly, with a loss that weights each position's TV by $B_{i-1}$ and equals the EAL deficit in expectation, is 1.9\% worse than TV when fine-tuning and 9.5\% from scratch (\cref{sec:appendix-causal}). 

Existing methods counter this attenuation with other position weights, such as DFlash's fixed decay, CAT's cumulative target-confidence weights, and D-PACE/D-PAKL's smoothed draft-confidence weights~\citep{DFlash,PARD2,DPACE}. These weights are chosen separately from the per-token loss. We instead derive the loss from EAL, giving a surrogate that removes the prefix products and strengthens undercoverage correction.

%% file: sections/method.tex
\section{TVP}
\label{sec:method}

\subsection{A Logarithmic Surrogate}

We first convert EAL to a sum, using a logarithm and Jensen's inequality,
\begin{equation}
\label{eq:tvp-jensen-zero}
\begin{aligned}
-\log\left(\frac{\EAL}{K}\right)
&\le -\frac1K\sum_{t=1}^{K}\E_p[\log B_t]\\
&=\E_p\sum_{i=1}^{K}w_iD_0(p_i,q_i)=L_0,
\end{aligned}
\end{equation}
where
\begin{equation}
\label{eq:tvp-dzero}
D_0(p,q)=
\sum_vp(v)\left[\log\frac{p(v)}{q(v)}\right]_+.
\end{equation}
The parameter gradient of $D_0(p,q)$ is
\[\nabla_\theta D_0
=-\sum_{v\in U}p(v)\nabla_\theta\log q(v),\]
and $\|\nabla_zD_0\|_1=2a(1-b)$.
This removes cross-position products and weights undercovered tokens by their target probabilities, resolving the vanishing gradient of TV. While the logit gradient is bounded, the loss itself can be unbounded. Consider a token with $p(v)=\epsilon$ and $q(v)=\epsilon e^{-M}$. The contribution to TV is $\epsilon(1-e^{-M})\le\epsilon$, while it contributes $\epsilon M$ to $D_0$. This does not imply that $D_0$ is unstable for training, but we seek a better-behaved loss function.

\subsection{The Power Family}

For better behavior, we want to soften $-\log r$, while modeling the cumulative dependence on earlier positions in the draft. A later position only matters if the tokens before it are accepted, so its penalty $\phi$ should reduce as the probability of accepting its prefix drops. For two successive factors $x$ and $y$ of a prefix product, we therefore ask that $\phi(xy) = \phi(x) + d(x)\phi(y)$, where $d(x) \in (0, 1]$ scales the later penalty by the survival $x$ of the prefix. Negative log has $d(x) = 1$ and ignores the prefix, while the rejection probability $1-xy=(1-x)+x(1-y)$ has $d(x)=x$. Appendix~\ref{app:tvp-composition} proves that this rule determines both $d$ and $\phi$. The penalty of three factors must not depend on whether they are grouped as $(xy)z$ or $x(yz)$, which implies that $d(xy) = d(x)d(y)$. For continuous $d$, we see that $d(x) = x^\beta$ for some $\beta \ge 0$.

Let us normalize $\phi(1) = 0$ and $\phi'(1) = -1$. Differentiating $\phi(xy) = \phi(x) + d(x)\phi(y)$ with respect to $y$ and then setting $y = 1$, we get $\phi'(x) = -x^{\beta-1}.$
Hence, 
\begin{equation}
\label{eq:tvp-phi}
\phi_\beta(r)=\frac{1-r^\beta}{\beta},
\qquad
\phi_0(r)=-\log r.
\end{equation}
Convexity, as needed for Jensen's inequality, requires $\beta \leq 1$, so we restrict $\beta \in [0, 1]$.

Applying $\phi_\beta$ to $r=\min\{1,q/p\}$ in place of $-\log r$, we define, for $0<\beta\le1$,
\begin{equation}
\label{eq:tvp-loss}
D_\beta(p,q)=\frac1\beta\sum_vp(v)
\left[
1-\min\left\{1,\frac{q(v)}{p(v)}\right\}^{\!\beta}
\right],
\end{equation}
and $L_\beta=\E_p\sum_iw_iD_\beta(p_i,q_i)$.
We have $D_\beta\to D_0$ as $\beta\to0$,
and $D_1=\TV$. Differentiation gives us
\begin{equation}
\label{eq:tvp-gradient}
\nabla_\theta D_\beta
=-\sum_{v\in U}
p(v)^{1-\beta}q(v)^\beta\nabla_\theta\log q(v).
\end{equation}
The coefficient interpolates between $p(v)$ at $\beta=0$
and $q(v)$ at $\beta=1$, and amplifies undercovered tokens over TV by a factor of
$(p(v)/q(v))^{1-\beta}$.

For $p(v)=\epsilon$, $q(v)=\epsilon e^{-M}$, the loss contribution is $\epsilon(1-e^{-\beta M})/\beta\le\epsilon/\beta$, and the score coefficient is $\epsilon e^{-\beta M}$. Thus, unlike $D_0$, any $\beta>0$ keeps the penalty on a badly undercovered token bounded.

\subsection{Annealing $\beta$}

As training progresses, we want to focus on acceptance more directly instead of undercoverage correction alone. We achieve this by annealing $\beta$ linearly from an initial value, typically 0, to 1, so that training moves from coverage to acceptance.

%% file: sections/theoretical_properties.tex
\section{Theoretical Properties}
\label{sec:tvp-properties}

We further motivate TVP by connecting it to EAL, showing that optimizing TVP guarantees a lower bound on EAL.

\begin{theorem}[EAL lower bound]
\label{thm:tvp-eal}
Let $h_i>0$ be position weights, $L_{\beta,h}=\E_p\sum_{i=1}^Kh_iD_\beta(p_i,q_i)$ and $C=\max_iw_i/h_i$. For $0<\beta\le1$,
\begin{equation}
\label{eq:tvp-eal-bound}
\EAL/K\ge[1-\beta CL_{\beta,h}]_+^{1/\beta}.
\end{equation}
At $\beta=0$, $\EAL/K\ge e^{-CL_{0,h}}$. With $h_i=w_i$, $C=1$ and $L_{\beta,h}=L_\beta$.
\end{theorem}
\begin{proof}[Proof sketch]
Applying Jensen's inequality to~\eqref{eq:tvp-eal} over the token trajectories, gives the first inequality:
\begin{equation*}
\begin{aligned}
\phi_\beta\left(\frac{\EAL}{K}\right)
&\le\frac1K\sum_{t=1}^K\E_p[\phi_\beta(B_t)]\\
&\le\E_p\sum_{i=1}^K w_iD_\beta(p_i,q_i)=L_\beta.
\end{aligned}
\end{equation*}
The second inequality follows from applying the composition rule repeatedly, then taking the conditional expectation over each token gives $D_\beta(p_i,q_i)$, and rearranging the sum gives $w_i$. Since $D_\beta\ge0$ and $w_i\le Ch_i$, $L_\beta\le CL_{\beta,h}$. For $\beta>0$, rearranging yields $(\EAL/K)^\beta\ge1-\beta L_\beta\ge1-\beta CL_{\beta,h}$. Combining this with $\EAL\ge0$ gives the positive-part bound. At zero, exponentiating the logarithmic inequality gives $\EAL/K\ge e^{-L_0}\ge e^{-CL_{0,h}}$.
\end{proof}

Every member of TVP therefore gives a lower bound on EAL, and $L_\beta\to0$ implies $\EAL\to K$ (Appendix~\ref{app:tvp-bound}).

We can also quantify the extra penalty that TVP places on undercoverage relative to TV.

\begin{proposition}[TV majorization]
\label{prop:tvp-majorization}
For $0<\beta\le1$,
\begin{equation}
\label{eq:tvp-majorization}
\TV(p,q)\le D_\beta(p,q)
\le\min\{D_0(p,q),\TV(p,q)/\beta\}.
\end{equation}
Moreover, $\KL(p\|q)\le D_0(p,q)\le\KL(p\|q)+\TV(p,q)$. $D_\beta$ is nonincreasing in $\beta$ and is zero exactly when $p=q$.
\end{proposition}

This ordering extends to the position-weighted sum $L_\beta$, which also bounds the EAL deficit.

\begin{proposition}[EAL-deficit majorization]
\label{prop:tvp-sequence-majorization}
For fixed target and draft distributions and $0\le\beta_1\le\beta_2\le1$,
\begin{equation}
\label{eq:tvp-monotone}
1-\EAL/K\le L_1\le L_{\beta_2}\le L_{\beta_1}.
\end{equation}
\end{proposition}

The proofs are given in \cref{app:tvp-majorization,app:tvp-gap}. Since every member of TVP upper bounds both the local rejection probability and the EAL deficit $1-\EAL/K$, lowering the loss at any point of the anneal raises the guaranteed lower bound on EAL. Training starts at $D_0$, which is within TV of KL and whose gradient does not vanish on undercovered tokens, and $D_\beta$ decreases to TV as $\beta\to1$.

Close losses do not imply close gradients, so we also bound how far the TVP logit gradient is from TV's.

\begin{theorem}[Recovery of TV's logit gradient]
\label{thm:tvp-gradient-convergence}
Let $g_\beta=\nabla_zD_\beta(p,\operatorname{softmax}(z))$. For $0\le\beta\le1$,
\begin{equation}
\label{eq:tvp-gradient-convergence}
\|g_\beta-g_1\|_1\le2(1-\beta)\TV(p,q).
\end{equation}
\end{theorem}

The extra corrective signal on undercovered tokens reduces as $\beta\to1$ for any $p$ and $q$, even when some draft probabilities are tiny, so the anneal ends with acceptance-focused updates. The gap also shrinks with $\TV(p,q)$, so even at low $\beta$, the coverage phase barely changes TV's update at positions the draft already predicts well, keeping the coverage penalty low. Appendix~\ref{app:tvp-annealing} gives the proof.

At $\beta=1$, TVP is an additive TV objective rather than the exact EAL deficit, so it keeps supervision on positions whose prefix is rarely accepted (\cref{app:tvp-gap}).

%% file: sections/experiments.tex
\section{Experiments}
\label{sec:experiments}

\subsection{Setup}
\label{sec:exp-setup}

\paragraph{Architecture.} Motivated by the recent success of parallel drafters, our experiments focus on DSpark and DFlash2 \citep{DSpark,inco2026dflash2}. We both train drafters from scratch and fine-tune released ones. We use the commonly reported targets Qwen3-8B, Qwen3-14B, and Qwen3.8-27B~\citep{Qwen3,qwen38}. For fine-tuning, we use DeepSeek-AI's released DSpark heads~\citep{deepspec_dspark_qwen3_8b,deepspec_dspark_qwen3_14b} and Qwen3-8B Eagle3 head~\citep{deepspec_eagle3_qwen3_8b}, and z-lab's Qwen3.8-27B DFlash2 head~\citep{zlab_qwen38_dflash2}.

\paragraph{Training.} We train on Open-PerfectBlend, minus any prompts in the evaluation sets, using thinking-on completions sampled from the target at $T=1$, with a fresh generation pass each epoch. Thinking stays on in evaluation too, since production reasoning models typically serve with it enabled, making this regime more predictive of deployed acceptance. The released heads may not have been trained on thinking-on data, so part of our gain over them may reflect this shift. To ensure fairness, every objective sees the exact same data in the same order, with the same optimizer, batch size, and learning-rate schedule. Unless noted, fine-tuning runs for one epoch at a peak learning rate of $2\times10^{-4}$, the best value of a sweep (\cref{sec:appendix-lr}), and pretraining uses $6\times10^{-4}$ for 5 epochs (DSpark) or 2 epochs (DFlash2). The main TVP schedule anneals $\beta$ linearly from 0 to 1 over the first 80\% of training. TVP with the linear weights $w_i$ from our derivation is within noise of using DSpark's exponential position weights ($-0.1\pm0.2\%$), so all per-token objectives share the exponential weights and differ only in their per-token loss. \Cref{thm:tvp-eal} covers them with $C\approx1.21$.

\paragraph{Baselines.} We compare against forward KL, target-token cross-entropy (CE), total variation (TV), $0.9$ TV + $0.1$ KL (TV+KL), and $0.9$ TV + $0.1$ CE (TV+CE), adapted from DSpark~\citep{DSpark}. For DSpark, we also compare against the published objectives CAT weighting~\citep{PARD2}, LK in its adaptive form and its log-acceptance form LK-log~\citep{LKLoss}, the acceptance-length objectives D-PACE and D-PAKL~\citep{DPACE}, DBLast-AL~\citep{DBLAST} and WTV~\citep{WTV}, and the block-verification loss BV~\citep{BV}. For DFlash2, we use the recipes of SpecForge~\citep{SpecForge} and TorchSpec~\citep{torchspec2026blog}. TVP variants with a fixed or later starting $\beta$ serve as ablations.

\paragraph{Evaluation.} We evaluate on nine benchmarks across math (GSM8K, MATH, AIME25), code (MBPP, HumanEval, LiveCodeBench) and chat (MT-Bench, Alpaca, Arena-Hard). Each drafter is served in SGLang~\citep{zheng2024sglang} with speculative sampling at $T=1$, and greedy decoding ($T=0$) is reported in \cref{sec:appendix-greedy}. We report $\tau$, the average number of tokens generated per target verification step, including the correction token. \Cref{sec:appendix-implementation} gives full hyperparameters and evaluation details.

$\Delta_{\mathrm{TV}}$ is the paired percentage difference against a TV objective trained in the same setting, with 95\% paired bootstrap confidence intervals (CIs) over prompts to cover evaluation noise. Training seed variance is discussed in \cref{sec:appendix-noise}. Bold marks the best entry in a column.

\subsection{Main Results}
\label{sec:exp-results}

\input{sections/tables/tab_main}
\input{sections/tables/tab_transfer}

\paragraph{TVP is the best objective.}
We first observe that on fine-tuning Qwen3-8B DSpark (\cref{tab:ft8b}), TVP achieves the highest $\tau$ among all objectives, outperforming TV by 1.5\%, KL by 8.2\% and CE by 9.8\%, and the released head by 25.7\%. On Qwen3-14B (\cref{tab:transfer}), we observe the same ordering and a similar margin over TV (+1.4\%), and on Eagle3, TVP is slightly ahead of TV (+0.2\%). For DFlash2, fine-tuning with TV or TVP improves the released head by 2.3 to 2.9\%, while the published CE recipes tie or degrade it (\cref{tab:dflash2-ft}). Pretraining once again shows that TVP maintains its advantage, with a 1.2\% improvement over the next best baseline (LK) and 4.2\% over TV (\cref{tab:pt8b}). Since DFlash2's path selector only considers the top-16 tokens, we train TV and TVP either on the top-16 tokens or on all tokens. Pretrained from scratch, TVP surpasses the best TV recipe by 1.3\% (\cref{tab:transfer-dflash2}). TVP gains 1.0\% from covering the full vocabulary, even though the selector only proposes the top-16 tokens, while restricting TV makes little difference, likely because TVP's coverage phase has little room in the top 16.

These gains hold beyond the default setup. Training with longer drafts makes the benefit of TVP more pronounced (+1.5\% to +1.9\% over TV, \cref{tab:drafts}), as later positions in a longer draft are harder to predict, so coverage matters more there. BV improves the most from $B=7$ to $B=15$, as it is a block-level objective. Likelihood objectives fall further behind TV at $B=15$ (KL from $-6.1\%$ to $-9.7\%$, CE from $-7.5\%$ to $-11.2\%$), so acceptance alignment matters more as drafts grow longer. TVP is also the best objective under block verification, surpassing even BV by 1.2\% at $B=7$ and 0.6\% at $B=15$. Its margin over TV grows under block verification, to +1.8\% at $B=7$ and +2.5\% at $B=15$, so TVP, a token-level acceptance objective, transfers well to a block verifier. TVP also stays ahead of TV under greedy decoding, by +1.6\% when fine-tuning and +3.6\% from scratch (\cref{sec:appendix-greedy}). TVP also has the highest $\tau$ at every learning rate we tried (\cref{tab:lr}), and the ranking of the objectives stays consistent from 1 to 5 epochs of training (\cref{tab:budget}), with TVP still ahead after 5 epochs, by +1.0\% when fine-tuning and +4.2\% from scratch, so its lead is not an artifact of short training. Finally, these margins are well above training-seed noise, as three seeds of TVP differ by at most 0.27\% (\cref{sec:appendix-noise}).

\input{sections/tables/tab_drafts}

\paragraph{Simple losses compete with published objectives.}
Interestingly, when fine-tuning (\cref{tab:ft8b}), none of the published baselines improves on plain TV beyond noise, with BV closest at +0.1\%. The baselines that come close to TV are the ones with a large component of TV in them (LK, TV+KL, TV+CE) and BV, which reduces to TV at the first position. Objectives that move further from per-token acceptance fall further behind. LK's log-acceptance form is 2.9\% below its adaptive form ($-3.2\%$ against $-0.3\%$), and D-PACE, which reweights CE by draft confidence, is 3.4\% below TV. Likelihood objectives fall furthest behind, with KL 6.1\% below TV on Qwen3-8B and 7.4\% below on Eagle3. At $T=1$, acceptance follows TV rather than likelihood. On both Qwen3-8B and Qwen3-14B, the KL-trained draft has the lowest forward KL in every entropy bucket, but up to $4\times$ TV's draft entropy, and its acceptance rate is 7\% lower than TV's on positions with target entropy below 1 nat, which make up 86\% of positions (\cref{tab:entropy-full}). This supports the notion that acceptance-oriented training is more effective than likelihood-oriented training when the model has already learned a reasonable policy. In pretraining, TV is still strong relative to pure likelihood objectives, but now falls short of every baseline that mixes in KL. However, simply adding a small KL term (TV+KL) brings it within noise of the best published baseline (+2.8\% against LK's +3.0\%). Prefix weighting does not help, as CAT applied to the same TV+KL loss is 0.6\% below it. The acceptance-length surrogates (D-PACE, D-PAKL, DBLast-AL and WTV) also do not beat TV beyond noise, consistent with the prefix attenuation described in \cref{sec:prefix-attenuation}. On DFlash2, both published CE recipes fall 4.2 to 9.9\% below TV. Across both regimes (\cref{tab:ft8b,tab:pt8b}), a simple loss matches or outperforms the newer objectives we tested, plain TV when fine-tuning and TV+KL from scratch.

\begin{figure*}[t]
  \centering
  \includegraphics[width=\textwidth]{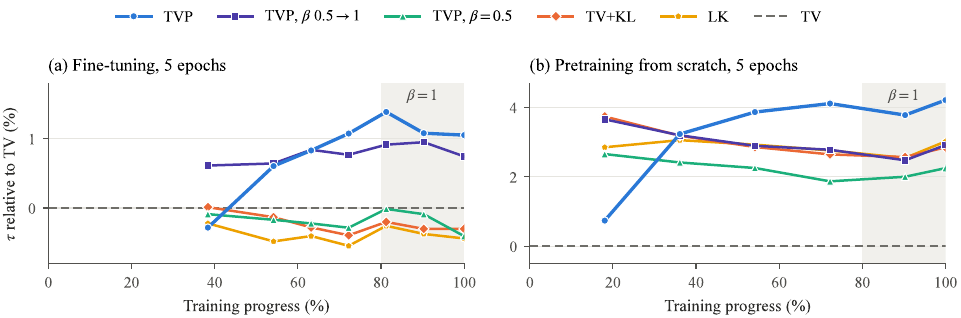}
  \caption{$\tau$ relative to TV at every evaluated checkpoint of the 5-epoch Qwen3-8B DSpark runs. TVP's $\beta$ reaches 1 at 80\% of training (shaded). \Cref{tab:curves} lists every checkpoint.}
  \label{fig:curves}
\end{figure*}

\input{sections/tables/tab_budget}
\input{sections/tables/tab_entropy}

\paragraph{Why TVP works.}
TVP's initial coverage-focused training is what allows it to surpass even pure acceptance-oriented training. Coverage alone is not enough, as TVP with $\beta$ fixed at 0 is 6.7\% below TV when fine-tuning (\cref{tab:ft8b}), close to KL. Fixing $\beta$ always underperforms annealing in both training regimes (\cref{tab:beta}), and starting the anneal from a higher $\beta$ also underperforms, especially in pretraining. From scratch, $\tau$ grows monotonically the earlier the anneal starts, from +2.3\% for a fixed $\beta=0.5$ to +2.9\%, +3.8\% and +4.2\% for starts at 0.5, 0.25 and 0 (\cref{tab:pt8b}), and Qwen3-14B shows the same order (\cref{tab:transfer}). This is because starting from $\beta=0$ gives undercovered tokens their strongest push (\cref{eq:tvp-gradient}), which helps most at positions with high target entropy (\cref{tab:entropy}), and annealing from a higher initial $\beta$ would skip it. The anneal's direction matters more than its shape, as reversing it costs 8.3\%, while ending the ramp at 40\% or stepping $\beta$ at 80\% costs 0.6\% and 1.3\% (\cref{tab:beta}).

The training curves in \cref{fig:curves} show the cost and payoff of this coverage phase. TVP starts below the anneal from $\beta=0.5$ and, when pretraining, below TV+KL and LK as well, but overtakes them while its $\beta$ is still below 1, by about 36\% of training from scratch and about 60 to 70\% when fine-tuning. The TV+KL, LK and fixed-$\beta$ objectives stay roughly parallel to TV for the whole run, while TVP's lead widens until $\beta$ reaches 1 and then stays roughly flat. Since TVP overtakes them before its loss becomes TV, the coverage phase sets up a better solution, rather than TVP simply ending on the right objective.

\Cref{tab:entropy} shows where the gain comes from, and Qwen3-8B shows the same pattern (\cref{tab:entropy-full}). On the lowest-entropy positions (48\% of positions), TVP and TV accept equally, and TVP's per-position gain over TV grows steadily with target entropy, to about $+0.02$. KL only leads above 2 nats, so TVP keeps part of KL's high-entropy advantage without its low-entropy cost. TVP's draft also has higher entropy and lower forward KL than TV's in every bucket, so the anneal yields a broader draft that covers more target mass.

%% file: sections/tables/tab_main.tex
\begin{table*}[t]
  \centering
  \begin{minipage}[t]{0.48\textwidth}
    \centering
    \caption{Qwen3-8B DSpark, fine-tuned for one epoch. Speedup is over the target alone at one request, and $\pm$ is the 95\% CI.}
    \label{tab:ft8b}
    \footnotesize
    \setlength{\tabcolsep}{4pt}
    \begin{tabular}{@{}lrrr@{}}
      \toprule
      Objective & $\tau$ & $\Delta_{\mathrm{TV}}$ (\%) & Speedup \\
      \midrule
      Released head & 3.566 & & \spd{2.36}{0.05} \\
      \midrule
      TV            & 4.415 & \tvref          & \spd{2.90}{0.09} \\
      KL            & 4.143 & $-6.1$\ci{0.2}  & \spd{2.70}{0.08} \\
      CE            & 4.083 & $-7.5$\ci{0.2}  & \spd{2.70}{0.09} \\
      TV+KL         & 4.404 & $-0.2$\ci{0.2}  & \spd{2.92}{0.09} \\
      TV+CE         & 4.410 & $-0.1$\ci{0.2}  & \spd{2.92}{0.09} \\
      \midrule
      LK            & 4.401 & $-0.3$\ci{0.3}  & \spd{2.87}{0.09} \\
      LK-log        & 4.273 & $-3.2$\ci{0.2}  & \spd{2.82}{0.09} \\
      D-PACE        & 4.266 & $-3.4$\ci{0.2}  & \spd{2.83}{0.09} \\
      WTV           & 4.390 & $-0.6$\ci{0.2}  & \spd{2.90}{0.09} \\
      BV            & 4.418 & $+0.1$\ci{0.2}  & \spd{2.86}{0.09} \\
      \midrule
      TVP, $\beta=0$  & 4.120 & $-6.7$\ci{0.2} & \spd{2.70}{0.08} \\
      TVP             & \textbf{4.482} & $+1.5$\ci{0.2} & \textbf{\spd{2.94}{0.09}} \\
      \bottomrule
    \end{tabular}
  \end{minipage}\hfill
  \begin{minipage}[t]{0.48\textwidth}
    \centering
    \caption{Qwen3-8B DSpark, pretrained from random initialization for five epochs.}
    \label{tab:pt8b}
    \footnotesize
    \setlength{\tabcolsep}{4pt}
    \begin{tabular}{@{}lrr@{}}
      \toprule
      Objective & $\tau$ & $\Delta_{\mathrm{TV}}$ (\%) \\
      \midrule
      TV    & 4.465 & \tvref         \\
      KL    & 4.334 & $-2.9$\ci{0.3} \\
      TV+KL & 4.592 & $+2.8$\ci{0.2} \\
      \midrule
      CAT       & 4.563 & $+2.2$\ci{0.2} \\
      LK        & 4.600 & $+3.0$\ci{0.2} \\
      D-PACE    & 4.436 & $-0.7$\ci{0.2} \\
      D-PAKL    & 4.479 & $+0.3$\ci{0.3} \\
      DBLast-AL & 4.426 & $-0.9$\ci{0.2} \\
      WTV       & 4.377 & $-2.0$\ci{0.3} \\
      \midrule
      TVP, $\beta=0.5$     & 4.565 & $+2.3$\ci{0.2} \\
      TVP, \anneal{0.5}    & 4.595 & $+2.9$\ci{0.2} \\
      TVP, \anneal{0.25}   & 4.634 & $+3.8$\ci{0.2} \\
      TVP                  & \textbf{4.653} & $+4.2$\ci{0.2} \\
      \bottomrule
    \end{tabular}
  \end{minipage}
\end{table*}

%% file: sections/tables/tab_transfer.tex
\begin{table}[t]
  \caption{Other targets and drafters. The 95\% CIs on $\Delta$ are $\pm0.2$, and $\pm0.4$ to $0.5$ for DFlash2.}
  \label{tab:transfer}
  \centering
  \footnotesize
  \setlength{\tabcolsep}{3pt}
  \newlength{\transferw}\settowidth{\transferw}{SpecForge CE}
  \begin{subtable}[c]{0.49\columnwidth}
    \centering
    \caption{Qwen3-14B DSpark, FT.}
    \label{tab:transfer-14b}
    \begin{tabular}{@{}lrr@{}}
      \toprule
      Objective & $\tau$ & $\Delta$ (\%) \\
      \midrule
      Released & 3.534 & \\
      \midrule
      TV        & 4.315 & \tvref \\
      KL        & 4.047 & $-6.2$ \\
      TV+KL     & 4.311 & $-0.1$ \\
      \midrule
      CAT       & 4.298 & $-0.4$ \\
      LK        & 4.307 & $-0.2$ \\
      D-PACE    & 4.170 & $-3.4$ \\
      D-PAKL    & 4.186 & $-3.0$ \\
      DBLast-AL & 4.197 & $-2.7$ \\
      WTV       & 4.285 & $-0.7$ \\
      \midrule
      TVP, $\beta=0.5$  & 4.307 & $-0.2$ \\
      TVP, \anneal{0.5} & 4.354 & $+0.9$ \\
      TVP               & \textbf{4.375} & $+1.4$ \\
      \bottomrule
    \end{tabular}
  \end{subtable}\hfill
  \begin{subtable}[c]{0.49\columnwidth}
    \centering
    \caption{Qwen3.8-27B DFlash2, PT.}
    \label{tab:transfer-dflash2}
    \begin{tabular}{@{}p{\transferw}rr@{}}
      \toprule
      Objective & $\tau$ & $\Delta$ (\%) \\
      \midrule
      Released & 3.741 & \\
      \midrule
      TV           & 3.541 & \tvref \\
      TV, top-16   & 3.555 & $+0.4$ \\
      SpecForge CE & 3.390 & $-4.2$ \\
      TorchSpec CE & 3.189 & $-9.9$ \\
      \midrule
      TVP, top-16  & 3.564 & $+0.6$ \\
      TVP          & \textbf{3.599} & $+1.7$ \\
      \bottomrule
    \end{tabular}

    \vspace{4pt}
    \caption{Qwen3-8B Eagle3, FT.}
    \label{tab:transfer-eagle3}
    \begin{tabular}{@{}p{\transferw}rr@{}}
      \toprule
      Objective & $\tau$ & $\Delta$ (\%) \\
      \midrule
      Released & 2.742 & \\
      \midrule
      TV    & 3.875 & \tvref \\
      KL    & 3.587 & $-7.4$ \\
      TV+KL & 3.846 & $-0.7$ \\
      \midrule
      TVP   & \textbf{3.881} & $+0.2$ \\
      \bottomrule
    \end{tabular}
  \end{subtable}
\end{table}

%% file: sections/tables/tab_drafts.tex
\begin{table}[t]
  \caption{Longer drafts and block verification~\citep{sun2025block}, Qwen3-8B DSpark FT. $\Delta$ is relative to TV under the same verifier and length, with 95\% CIs of $\pm0.2$ ($B=7$) and $\pm0.3$ ($B=15$).}
  \label{tab:drafts}
  \centering
  \footnotesize
  \setlength{\tabcolsep}{3.2pt}
  \begin{tabular}{@{}lrrrrrr@{}}
    \toprule
    & \multicolumn{2}{c}{Token, $B=15$} & \multicolumn{2}{c}{Block, $B=7$} & \multicolumn{2}{c}{Block, $B=15$} \\
    \cmidrule(lr){2-3} \cmidrule(lr){4-5} \cmidrule(l){6-7}
    Objective & $\tau$ & $\Delta$ (\%) & $\tau$ & $\Delta$ (\%) & $\tau$ & $\Delta$ (\%) \\
    \midrule
    Released head & 3.571 & & 3.667 & & 3.668 & \\
    \midrule
    TV            & 4.893 & \tvref  & 4.533 & \tvref & 5.096 & \tvref \\
    KL            & 4.417 & $-9.7$  & 4.294 & $-5.3$ & 4.636 & $-9.0$ \\
    CE            & 4.343 & $-11.2$ & 4.232 & $-6.6$ & 4.558 & $-10.5$ \\
    TV+KL         & 4.881 & $-0.2$  & 4.554 & $+0.5$ & 5.131 & $+0.7$ \\
    TV+CE         & 4.854 & $-0.8$  & 4.559 & $+0.6$ & 5.121 & $+0.5$ \\
    \midrule
    LK            & 4.837 & $-1.1$  & 4.547 & $+0.3$ & 5.084 & $-0.2$ \\
    LK-log        & 4.604 & $-5.9$  & 4.424 & $-2.4$ & 4.829 & $-5.2$ \\
    WTV           & 4.842 & $-1.0$  & 4.503 & $-0.7$ & 5.052 & $-0.9$ \\
    BV            & 4.944 & $+1.0$  & 4.557 & $+0.5$ & 5.194 & $+1.9$ \\
    \midrule
    TVP           & \textbf{4.987} & $+1.9$ & \textbf{4.613} & $+1.8$ & \textbf{5.225} & $+2.5$ \\
    \bottomrule
  \end{tabular}
\end{table}

%% file: sections/tables/tab_budget.tex
\begin{table}[t]
  \caption{Training budget, with learning rates $10^{-4}$ (FT) and $6\times10^{-4}$ (PT).}
  \label{tab:budget}
  \centering
  \footnotesize
  \setlength{\tabcolsep}{4pt}
  \begin{tabular}{@{}lrrrr@{}}
    \toprule
    & \multicolumn{2}{c}{FT epochs} & \multicolumn{2}{c}{PT epochs} \\
    \cmidrule(lr){2-3} \cmidrule(l){4-5}
    Objective & 1 & 5 & 2 & 5 \\
    \midrule
    TV & 4.399 & 4.633 & 4.217 & 4.465 \\
    KL & 4.100 & 4.343 & 4.082 & 4.334 \\
    \midrule
    CAT    & 4.371 & 4.614 & 4.335 & 4.563 \\
    D-PACE & 4.244 & 4.480 & 4.205 & 4.436 \\
    D-PAKL & 4.242 & 4.497 & 4.242 & 4.479 \\
    \midrule
    TVP, $\beta=0.5$  & 4.370 & 4.614 & 4.330 & 4.565 \\
    TVP, \anneal{0.5} & 4.423 & 4.667 & 4.364 & 4.595 \\
    TVP               & \textbf{4.437} & \textbf{4.681} & \textbf{4.417} & \textbf{4.653} \\
    \midrule
    Spearman $\rho$ & \multicolumn{2}{c}{0.95} & \multicolumn{2}{c}{0.98} \\
    \bottomrule
  \end{tabular}
\end{table}

%% file: sections/tables/tab_entropy.tex
\begin{table}[!tb]
  \caption{Acceptance, draft entropy and forward KL by target entropy (nats), Qwen3-14B FT, last 5\% of training. Acceptance is the mean of $1-\TV$. Underlined: within noise (0.01) of the best. \Cref{tab:entropy-full} adds all buckets and Qwen3-8B.}
  \label{tab:entropy}
  \centering
  \footnotesize
  \setlength{\tabcolsep}{1.9pt}
  \begin{tabular}{@{}lrccccccccc@{}}
    \toprule
    & & \multicolumn{3}{c}{Acceptance} & \multicolumn{3}{c}{Draft entropy} & \multicolumn{3}{c}{Forward KL} \\
    \cmidrule(lr){3-5} \cmidrule(lr){6-8} \cmidrule(l){9-11}
    Target & Share & TVP & TV & KL & TVP & TV & KL & TVP & TV & KL \\
    \midrule
    $<0.05$ & 47.6\% & \underline{.876} & \textbf{.877} & .819 & 0.19 & 0.16 & 0.64 & 0.66 & 0.79 & 0.46 \\
    0.2--0.5 & 10.9\% & \textbf{.636} & \underline{.631} & .569 & 0.67 & 0.59 & 1.60 & 1.73 & 2.06 & 1.16 \\
    1--2    & 13.0\% & \underline{.474} & .460 & \textbf{.479} & 1.27 & 1.17 & 2.36 & 2.48 & 2.95 & 1.51 \\
    $>3$    & 0.1\%  & .257 & .239 & \textbf{.312} & 2.58 & 2.39 & 4.51 & 4.14 & 4.75 & 2.47 \\
    \bottomrule
  \end{tabular}
\end{table}

%% file: sections/conclusion.tex
\section{Conclusions}
\label{sec:conclusion}
\label{sec:discussion}
\label{sec:limitations}

Acceptance-focused training matters most once a drafter already has a good policy. In our main fine-tuning comparison, plain TV matches or outperforms every published objective we tested, including those built around acceptance length, and likelihood objectives trail it by 6 to 8\%. A well-tuned TV baseline should therefore accompany new draft losses.

Coverage matters when training from scratch. A randomly initialized drafter spreads its mass over the whole vocabulary and puts almost none on the tokens the target favors, which is exactly where TV's gradient vanishes. TV hence falls 2.2 to 3.0\% behind its mixtures with KL. A simple mixture of 90\% TV and 10\% KL recovers two thirds of TVP's gain over TV ($+2.8\%$ against $+4.2\%$), while optimizing EAL directly is 9.5\% worse than TV, because prefix products starve later positions.

TVP blends coverage and acceptance as a surrogate for EAL, and its two phases interpolate smoothly with $\beta$. Annealing $\beta$ from zero to one gives the highest or tied-highest $\tau$ in every sampling setting we evaluated, including longer drafts and block verification, where it outperforms BV, a loss designed for block verification, despite being a token-level loss.

\paragraph{Limitations.}
Our Eagle3 runs see far fewer training samples than our DSpark runs, so the near tie between TVP and TV there ($+0.2\%$) may not reflect a fully trained Eagle3 head. TVP ramps up more slowly, trailing TV+KL and LK for about the first third of pretraining (\cref{fig:curves}), which could matter under a low training budget. The anneal also assumes that the training budget is known in advance, and adaptive schedules remain open. We evaluate published objectives within a shared drafter and training recipe, so our comparisons isolate the loss rather than reproduce each paper's full system. We also only evaluate at $T=1$ and with greedy decoding, not at intermediate temperatures, and measure speedups at a fixed concurrency that does not represent a production workload.

%% file: sections/acknowledgements.tex
\section*{Acknowledgements}
The author thanks Joyjit Daw and Phil Howes for their support of this work, Max Kirkby for his review and valuable feedback on the manuscript, and all three for their encouragement.

%% file: sections/impact_statement.tex
\section*{Impact Statement}
This paper studies how to train draft models for speculative decoding. Better drafters reduce the cost and energy of inference. Since speculative sampling is lossless, our work inherits the societal impacts of the models it accelerates and adds no new capabilities. Cheaper inference can broaden access to large language models, which can amplify both their beneficial and harmful uses.

%% file: sections/references.tex
\bibliographystyle{icml2026}
\bibliography{references}

%% file: sections/appendix/appendix.tex
\input{sections/appendix/appendix_tvp}

\input{sections/appendix/appendix_loss}
\input{sections/appendix/appendix_experiments}
\input{sections/appendix/appendix_results}

%% file: sections/appendix/appendix_tvp.tex
\section{TVP Identities and Proofs}
\label{app:tvp-proofs}

\Cref{app:tvp-eal} proves the EAL identity~\eqref{eq:tvp-eal} and the attenuation bound~\eqref{eq:tvp-attenuation} from \cref{sec:background}. \Cref{app:tvp-gradients} proves the gradient formulas~\eqref{eq:tvp-tv-kl-grad}, \eqref{eq:tvp-tv-kl-norm} and~\eqref{eq:tvp-gradient}, and \cref{app:tvp-composition} derives the power family~\eqref{eq:tvp-phi}. \Cref{app:tvp-bound,app:tvp-majorization,app:tvp-annealing} prove the results of \cref{sec:tvp-properties}.

\subsection{EAL Identities and Gradients}
\label{app:tvp-eal}

We use notation from Sections~\ref{sec:background} and \ref{sec:method}, and follow the convention of conditioning on a fixed context throughout. We write $[u]_+=\max\{u,0\}$ for the positive part, and square brackets around a condition, such as $[q_i(Y_i)<p_i(Y_i)]$, for its indicator. For derivatives, we fix the target distribution and assume the probabilities are positive. Where $q(v)=p(v)$, the positive part is not differentiable, and we take its derivative to be zero there, for every $\beta$ and in every gradient identity below. Probability bounds extend to zero probabilities by limits, allowing $D_0=+\infty$.

\paragraph{Change of measure and EAL.}
For a particular prefix $y_{1:t}$, its probability of being proposed and fully accepted is
\begin{align*}
q(y_{1:t}\mid x)A_t(y_{1:t})
&=\prod_{i=1}^t q_i(y_i)\min\left\{1,\frac{p_i(y_i)}{q_i(y_i)}\right\}\\
&=\prod_{i=1}^t\min\{p_i(y_i),q_i(y_i)\}\\
&=\prod_{i=1}^t p_i(y_i)\min\left\{1,\frac{q_i(y_i)}{p_i(y_i)}\right\}\\
&=p(y_{1:t}\mid x)B_t(y_{1:t}).
\end{align*}
Summing over all prefixes of length $t$ gives $\Pr(N\ge t)=\E_q[A_t]=\E_p[B_t]$. Thus,
\begin{equation*}
\EAL=\sum_{t=1}^K\Pr(N\ge t)=\sum_{t=1}^K\E_q[A_t]=\sum_{t=1}^K\E_p[B_t],
\end{equation*}
which proves~\eqref{eq:tvp-eal}.

\paragraph{Exact TV representation.}
Observe that $B_{i-1}-B_i=B_{i-1}(1-r_i)$. Summing these from $i=1$ to $t$ gives
\begin{equation*}
1-B_t=\sum_{i=1}^t B_{i-1}(1-r_i).
\end{equation*}
For a fixed prefix $Y_{<i}$, the next token is drawn from $p_i$. Therefore,
\begin{align*}
\E_p[1-r_i\mid Y_{<i}]
&=\sum_{v\in\mathcal V}p_i(v)\left(1-\min\left\{1,\frac{q_i(v)}{p_i(v)}\right\}\right)\\
&=\sum_{v\in\mathcal V}[p_i(v)-q_i(v)]_+\\
&=\TV(p_i,q_i).
\end{align*}
$B_{i-1}$ depends only on $Y_{<i}$, so we can take it outside the conditional expectation. We can then compute the following:
\begin{align*}
1-\frac{\EAL}{K}
&=\frac{1}{K}\sum_{t=1}^K\E_p[1-B_t]\\
&=\frac{1}{K}\sum_{t=1}^K\sum_{i=1}^t\E_p\!\left[B_{i-1}\E_p[1-r_i\mid Y_{<i}]\right]\\
&=\frac{1}{K}\sum_{i=1}^K\sum_{t=i}^K\E_p[B_{i-1}\TV(p_i,q_i)].
\end{align*}
Position $i$ appears in exactly $K-i+1$ horizons. Using the weights $w_i=(K-i+1)/K$, we get
\begin{equation}
\label{eq:tvp-exact-tv-appendix}
1-\frac{\EAL}{K}=\E_p\!\left[\sum_{i=1}^K w_iB_{i-1}\TV(p_i,q_i)\right].
\end{equation}

\paragraph{Exact EAL gradient.}
When $q_i(Y_i)<p_i(Y_i)$, $\nabla_\theta r_i=r_i\nabla_\theta\log q_i(Y_i)$ and when $q_i(Y_i)>p_i(Y_i)$, $\nabla_\theta r_i = 0$. Hence
\begin{align*}
\nabla_\theta B_t
&=B_t\sum_{i=1}^t[q_i(Y_i)<p_i(Y_i)]\nabla_\theta\log q_i(Y_i).
\end{align*}
The target distribution $p$ does not depend on $\theta$, so we can differentiate through $\E_p$ with respect to $\theta$ to see that
\begin{align*}
\nabla_\theta\!\left(\frac{\EAL}{K}\right)
&=\E_p\!\left[\frac{1}{K}\sum_{t=1}^K\sum_{i=1}^t B_t[q_i(Y_i)<p_i(Y_i)]\nabla_\theta\log q_i(Y_i)\right]\\
&=\E_p\!\left[\sum_{i=1}^K\left(\frac{1}{K}\sum_{t=i}^K B_t\right)[q_i(Y_i)<p_i(Y_i)]\nabla_\theta\log q_i(Y_i)\right]\\
&=\E_p\!\left[\sum_{i=1}^K c_i[q_i(Y_i)<p_i(Y_i)]\nabla_\theta\log q_i(Y_i)\right].
\end{align*}
Here $c_i=\frac1K\sum_{t=i}^KB_t$. For $t\ge i$, every additional factor in $B_t$ lies in $[0,1]$, so $B_t\le B_i$. Averaging over the $K-i+1$ remaining horizons gives $c_i\le w_iB_i$, the bound stated after~\eqref{eq:tvp-attenuation}.

\paragraph{TVP gradient.}
For the TVP objective $L_\beta=\E_p\sum_{i=1}^K w_i D_\beta(p_i,q_i)$, conditioning on $Y_{<i}$ gives $D_\beta(p_i,q_i)=\E_p[\phi_\beta(r_i)\mid Y_{<i}]$. Since $\phi_\beta'(r)=-r^{\beta-1}$, including $\beta=0$, differentiating each $\phi_\beta(r_i)$ gives
\begin{equation}
\nabla_\theta L_\beta=-\E_p\!\left[\sum_{i=1}^K w_i r_i^\beta[q_i(Y_i)<p_i(Y_i)]\nabla_\theta\log q_i(Y_i)\right].
\end{equation}
These score coefficients retain the positional weights and the local factor $r_i^\beta$, providing supervision independently of the survival of earlier positions.

\subsection{Gradient Formulas and Norm Bounds}
\label{app:tvp-gradients}

\paragraph{Score gradients.}
Letting $U=\{v:p(v)>q(v)\}$ we see that,
\begin{align*}
\nabla_\theta\TV(p,q)
&=-\sum_{v\in U}\nabla_\theta q(v)
=-\sum_{v\in U}q(v)\nabla_\theta\log q(v),\\
\nabla_\theta\KL(p\|q)&=-\sum_v p(v)\nabla_\theta\log q(v).
\end{align*}
This shows \eqref{eq:tvp-tv-kl-grad}. For $\beta>0$, only tokens in $U$ contribute to \eqref{eq:tvp-loss}, giving
\begin{equation*}
D_\beta(p,q)=\frac{1}{\beta}\sum_{v\in U}\bigl[p(v)-p(v)^{1-\beta}q(v)^\beta\bigr].
\end{equation*}
Differentiating each term gives
\begin{align*}
\nabla_\theta D_\beta(p,q)
&=-\sum_{v\in U}p(v)^{1-\beta}q(v)^{\beta-1}\nabla_\theta q(v)\\
&=-\sum_{v\in U}p(v)^{1-\beta}q(v)^\beta\nabla_\theta\log q(v),
\end{align*}
proving \eqref{eq:tvp-gradient}. Differentiation of $D_0$ gives the same expression with $\beta=0$. Since $q(v)\le p(v)$ on $U$, the coefficient satisfies $q(v)\le p(v)^{1-\beta}q(v)^\beta\le p(v)$.

\paragraph{Logit gradient L1 norms.}
Recall that $U=\{v:q(v)<p(v)\}$ is the set of undercovered tokens, and $a=p(U)$ and $b=q(U)$ are their target and draft mass. For each token $v$, let $a_{\beta,v}=p(v)^{1-\beta}q(v)^\beta$ if $v\in U$ and $a_{\beta,v}=0$ otherwise. Let $a_\beta$ be the vector with entries $a_{\beta,v}$, and $A_\beta=\sum_v a_{\beta,v}$ its total. Writing $e_v$ for the one-hot vector of token $v$, the softmax gives $\nabla_z\log q(v)=e_v-q$, so
\begin{equation}
\label{eq:tvp-logit-vector}
g_\beta=\nabla_zD_\beta=-\sum_v a_{\beta,v}(e_v-q)=A_\beta q-a_\beta.
\end{equation}
Since $q(v)\le a_{\beta,v}\le p(v)$ on $U$, we have $b\le A_\beta\le a\le1$. Hence $g_{\beta,v}=A_\beta q(v)-a_{\beta,v}\le0$ on $U$, while $g_{\beta,v}=A_\beta q(v)\ge0$ outside $U$. Thus, the sum of positive coordinates of $g_\beta$ is
\begin{equation*}
\sum_{v\notin U}g_{\beta,v}=A_\beta\sum_{v\notin U}q(v)=A_\beta(1-b).
\end{equation*}
The negative terms have the same total magnitude. Therefore,
\begin{equation*}
\|g_\beta\|_1=2A_\beta(1-b).
\end{equation*}
Taking $\beta=1$ and $\beta=0$ gives $2b(1-b)$ and $2a(1-b)$, respectively. For KL, $\nabla_z\KL=q-p$, whose norm is $2\TV(p,q)=2(a-b)$. This establishes~\eqref{eq:tvp-tv-kl-norm} and places TVP's logit gradient norm between those of TV and $D_0$.

\subsection{The Power Composition Rule}
\label{app:tvp-composition}

\begin{lemma}[Power family under a composition rule]
\label{lem:tvp-composition}
Let $\phi:(0,1]\to[0,\infty)$ be continuously differentiable, with $\phi(1)=0$ and left derivative $\phi'(1)=-1$. Let $d:(0,1]\to(0,1]$ be continuous. If $\phi(xy)=\phi(x)+d(x)\phi(y)$ for all $x,y\in(0,1]$, then $d(x)=x^\beta$ for some $\beta\ge0$, and
\begin{equation*}
\phi(x)=
\begin{cases}
\dfrac{1-x^\beta}{\beta}, & \beta>0,\\[4pt]
-\log x, & \beta=0.
\end{cases}
\end{equation*}
Within this family, boundedness at zero and convexity hold together exactly when $0<\beta\le1$.
\end{lemma}
\begin{proof}
Applying the composition rule to the groupings $(xy)z$ and $x(yz)$, we see that
\begin{align*}
\phi((xy)z)&=\phi(xy)+d(xy)\phi(z)=\phi(x)+d(x)\phi(y)+d(xy)\phi(z),\\
\phi(x(yz))&=\phi(x)+d(x)\phi(yz)=\phi(x)+d(x)\phi(y)+d(x)d(y)\phi(z).
\end{align*}
Thus $[d(xy)-d(x)d(y)]\phi(z)=0$. Since $\phi'(1)=-1$ ensures that $\phi$ is not identically zero, we can choose a $z$ such that $\phi(z) > 0$ giving us $d(xy)=d(x)d(y)$ for all $x,y\in(0,1]$.

Let us define $f(t)=-\log d(e^{-t})$ for $t\ge0$. The multiplicativity of $d$ implies $f(s+t)=f(s)+f(t)$. $f$ is continuous because of the continuity of $d$. Let $\beta=f(1)$. Then, additivity gives $f(m/n)=\beta m/n$ for nonnegative integers $m$ and positive integers $n$. Continuity implies $f(t) = \beta t$ for all real $t\ge0$. Since $d\le1$, we have $f\ge0$ and hence $\beta\ge0$. Letting $t=-\log x$ then gives $d(x)=x^\beta$.

Differentiating the composition rule in $y$ from the left at $y=1$ yields
\begin{equation*}
x\phi'(x)=d(x)\phi'(1)=-x^\beta,
\qquad
\phi'(x)=-x^{\beta-1}.
\end{equation*}
Integrating from $x$ to $1$ and using $\phi(1)=0$ gives
\begin{equation*}
\phi(x)=\int_x^1 t^{\beta-1}\,dt
=
\begin{cases}
\dfrac{1-x^\beta}{\beta}, & \beta>0,\\[4pt]
-\log x, & \beta=0.
\end{cases}
\end{equation*}
For $\beta>0$, the limit at zero is $\phi(0+)=\frac{1}{\beta}$, while the logarithmic endpoint diverges. Finally, $\phi''(x)=(1-\beta)x^{\beta-2}$, which is nonnegative exactly when $\beta\le1$. Both boundedness and convexity therefore hold iff $0<\beta\le1$.
\end{proof}
Since $\frac{1-x^\beta}{\beta}\to-\log x$ as $\beta\downarrow0$, the logarithmic member is the limit of the power family, so $D_\beta\to D_0$ as $\beta\to0$.

\subsection{Proof of the EAL Lower Bound}
\label{app:tvp-bound}

\begin{proof}[Proof of Theorem~\ref{thm:tvp-eal}]
Repeatedly applying the composition rule gives
\begin{equation}
\label{eq:tvp-prefix-power}
\phi_\beta(B_t)=\sum_{i=1}^t B_{i-1}^\beta\phi_\beta(r_i)\le\sum_{i=1}^t\phi_\beta(r_i),
\end{equation}
because $B_{i-1}^\beta\le1$ and each $\phi_\beta(r_i)$ is nonnegative. For $\beta=0$, we get back the exact logarithmic additivity. Taking the conditional expectation of each local penalty gives
\begin{equation*}
\E_p[\phi_\beta(r_i)\mid Y_{<i}]
=\sum_{v\in\mathcal V}p_i(v)\phi_\beta\!\left(\min\left\{1,\frac{q_i(v)}{p_i(v)}\right\}\right)
=D_\beta(p_i,q_i).
\end{equation*}
As $\phi_\beta$ is convex for $0\le\beta\le1$, we can apply Jensen's inequality, with \eqref{eq:tvp-prefix-power}, to give
\begin{align*}
\phi_\beta\!\left(\frac{\EAL}{K}\right)
&=\phi_\beta\!\left(\frac{1}{K}\sum_{t=1}^K\E_p[B_t]\right)\\
&\le\frac{1}{K}\sum_{t=1}^K\E_p[\phi_\beta(B_t)]\\
&\le\frac{1}{K}\sum_{t=1}^K\sum_{i=1}^t\E_p\bigl[\E_p[\phi_\beta(r_i)\mid Y_{<i}]\bigr]\\
&=\sum_{i=1}^K\frac{K-i+1}{K}\E_p[D_\beta(p_i,q_i)]\\
&=L_\beta.
\end{align*}
For $\beta>0$, substituting $\phi_\beta$ and rearranging gives
\begin{equation*}
\left(\frac{\EAL}{K}\right)^\beta\ge1-\beta L_\beta.
\end{equation*}
The left-hand side is also nonnegative. Taking the $\frac{1}{\beta}$ power therefore gives $\frac{\EAL}{K}\ge[1-\beta L_\beta]_+^{1/\beta}$. At $\beta=0$, the same argument gives $-\log\!\left(\frac{\EAL}{K}\right)\le L_0$, and exponentiating yields $\frac{\EAL}{K}\ge e^{-L_0}$. Finally, for general weights $h_i$, $D_\beta\ge0$ and $w_i\le Ch_i$ give $L_\beta\le CL_{\beta,h}$. Both bounds are nonincreasing in $L_\beta$, so they also hold with $CL_{\beta,h}$ in place of $L_\beta$.
\end{proof}

For DSpark's weights $h_i=\exp(-(i-1)/4)$ and $K=7$, the largest ratio is at $i=4$, giving $C=(4/7)e^{3/4}\approx1.21$. The trained loss also divides by $\sum_ih_i$, which only rescales $L_{\beta,h}$ by a constant.

\paragraph{Prefix-survival bounds.}
The same argument can be applied to a single $B_t$ term, giving us for $\beta>0$,
\begin{equation}
\label{eq:tvp-prefix-bound}
\Pr(N\ge t)\ge\left[1-\beta\sum_{i=1}^t\E_p[D_\beta(p_i,q_i)]\right]_+^{1/\beta}.
\end{equation}
At $\beta=0$, the right-hand side becomes $\exp\!\left(-\sum_{i=1}^t\E_p[D_0(p_i,q_i)]\right)$. Summing these bounds over $t$ gives a lower bound on EAL that is at least as strong as the aggregated bound in Theorem~\ref{thm:tvp-eal}. This is because the average of $\sum_{i=1}^t\E_p[D_\beta(p_i,q_i)]$ over $t=1,\dots,K$ is $L_\beta$, and $u\mapsto[1-\beta u]_+^{1/\beta}$ is convex for $0<\beta\le1$. Jensen's inequality hence bounds the average of the prefix lower bounds from below by $[1-\beta L_\beta]_+^{1/\beta}$. The same argument applies to $u\mapsto e^{-u}$ at $\beta=0$.

\subsection{Majorization and the EAL Gap}
\label{app:tvp-majorization}

\begin{proof}[Proof of Proposition~\ref{prop:tvp-majorization}]
For $0<r\le1$ and $0<\beta\le1$, the integral representation from Lemma~\ref{lem:tvp-composition} gives
\begin{equation*}
1-r=\int_r^1 1\,dt
\le\int_r^1 t^{\beta-1}\,dt
=\phi_\beta(r)
\le\int_r^1\frac{1}{t}\,dt=-\log r.
\end{equation*}
Also $r^\beta\ge r$, so $\phi_\beta(r)\le\frac{1-r}{\beta}$. Substituting $r = \min\{1, q/p\}$ gives us $\E_p[1-r] = \TV(p,q)$. Hence
\begin{equation*}
\TV(p,q)\le D_\beta(p,q)\le\min\left\{D_0(p,q),\frac{\TV(p,q)}{\beta}\right\},
\end{equation*}
which proves~\eqref{eq:tvp-majorization}.

For the bounds on $D_0$, note that $D_0$ keeps only the positive terms of $\KL(p\|q)$, so
\begin{equation*}
D_0(p,q)-\KL(p\|q)=\sum_{v:\,q(v)>p(v)}p(v)\log\frac{q(v)}{p(v)}\ge0.
\end{equation*}
Applying $\log x\le x-1$ to each term bounds this sum by $\sum_{v:\,q(v)>p(v)}\bigl(q(v)-p(v)\bigr)=\TV(p,q)$, so $D_0(p,q)\le\KL(p\|q)+\TV(p,q)$.

For each $t\in(0,1]$, the integrand $t^{\beta-1}$ is nonincreasing in $\beta$. The same is therefore true of $\phi_\beta(r)$ and its expectation $D_\beta(p,q)$. If $D_\beta(p,q)=0$, the lower bound forces $\TV(p,q)=0$, so $p=q$. Conversely, $p=q$ makes every $\phi_\beta(r_i)$ zero.
\end{proof}
Subtracting TV from the upper and lower bounds also gives
\begin{equation}
0\le D_\beta-\TV\le\frac{1-\beta}{\beta}\TV,
\qquad 0<\beta\le1.
\end{equation}
Multiplying by the positional weights and averaging gives $0\le L_\beta-L_1\le\frac{1-\beta}{\beta}L_1$.

\subsubsection{Exact gap decomposition}
\label{app:tvp-gap}
Substituting equation~\eqref{eq:tvp-exact-tv-appendix} above gives
\begin{equation}
\label{eq:tvp-gap}
\begin{aligned}
L_\beta-\left(1-\frac{\EAL}{K}\right)
&=\E_p\!\left[\sum_{i=1}^K w_i\bigl(D_\beta(p_i,q_i)-B_{i-1}\TV(p_i,q_i)\bigr)\right]\\
&=\E_p\!\left[\sum_{i=1}^K w_i\bigl(D_\beta(p_i,q_i)-\TV(p_i,q_i)\bigr)\right]+\E_p\!\left[\sum_{i=1}^K w_i(1-B_{i-1})\TV(p_i,q_i)\right].
\end{aligned}
\end{equation}

Both terms are nonnegative. The first measures how far $D_\beta$ exceeds TV and decreases to zero as $\beta$ increases to one. The second measures the supervision added by removing prefix-survival weights. At $\beta=1$, only the second term remains.

\begin{proof}[Proof of Proposition~\ref{prop:tvp-sequence-majorization}]
Setting $\beta=1$ in~\eqref{eq:tvp-gap} leaves
\begin{equation*}
L_1-\left(1-\frac{\EAL}{K}\right)
=\E_p\!\left[\sum_{i=1}^K w_i(1-B_{i-1})\TV(p_i,q_i)\right]\ge0.
\end{equation*}
For $0\le\beta_1\le\beta_2\le1$, local majorization gives $D_1(p_i,q_i)\le D_{\beta_2}(p_i,q_i)\le D_{\beta_1}(p_i,q_i)$ at every target prefix. Multiplying by the nonnegative weights and averaging under the fixed target distribution preserves these inequalities. Combining them with the displayed identity yields
\begin{equation*}
1-\frac{\EAL}{K}\le L_1\le L_{\beta_2}\le L_{\beta_1}.
\end{equation*}
The logarithmic endpoint follows by taking the limit as $\beta\downarrow0$.
\end{proof}

\subsection{Gradient Convergence and Annealing}
\label{app:tvp-annealing}

\begin{proof}[Proof of Theorem~\ref{thm:tvp-gradient-convergence}]
Let $\Delta a_v=a_{\beta,v}-a_{1,v}$, with $a_{\beta,v}$ as in~\eqref{eq:tvp-logit-vector}, and let $\Delta a$ be the vector with entries $\Delta a_v$. On $U$, $\Delta a_v=p(v)\bigl[(q(v)/p(v))^\beta-q(v)/p(v)\bigr]\ge0$, since $r^\beta\ge r$ for $r\in(0,1]$, and $\Delta a_v=0$ outside $U$. By~\eqref{eq:tvp-logit-vector}, $g_\beta-g_1=\bigl(\sum_v\Delta a_v\bigr)q-\Delta a$, and using $\|q\|_1=1$ gives
\begin{align*}
\|g_\beta-g_1\|_1
&=\left\|\left(\sum_v\Delta a_v\right)q-\Delta a\right\|_1\\
&\le\left(\sum_v\Delta a_v\right)\|q\|_1+\|\Delta a\|_1\\
&=2\sum_v\Delta a_v.
\end{align*}
For $r\in(0,1]$, concavity of $r\mapsto r^\beta$ bounds it by its tangent at $1$, giving $r^\beta\le1-\beta+\beta r$. Thus $r^\beta-r\le(1-\beta)(1-r)$, and
\begin{align*}
\sum_v\Delta a_v
&=\sum_{v\in U}p(v)\left[\left(\frac{q(v)}{p(v)}\right)^\beta-\frac{q(v)}{p(v)}\right]\\
&\le(1-\beta)\sum_{v\in U}[p(v)-q(v)]\\
&=(1-\beta)\TV(p,q).
\end{align*}
Combining the inequalities proves $\|g_\beta-g_1\|_1\le2(1-\beta)\TV(p,q)$. Since $\TV(p,q)\le1$, the convergence is uniform.
\end{proof}

%% file: sections/appendix/appendix_loss.tex
\section{Training Objectives and Baseline Definitions}
\label{sec:appendix-losses}

We define every objective in our comparisons. Each published loss follows its paper and is trained with the same drafter, data and optimizer as the other objectives. Where a paper also proposes an architecture or a different base loss (PARD-2, DBLast), we take only its loss or weighting, as noted below. Unless specified otherwise, $p_i$ and $q_i$ are the target and draft distributions at position $i$ of a training block, and $y_i$ is the target-sampled token. A block contains $K$ prediction positions, excluding the anchor token. We use $[z]_+=\max\{0,z\}$ and $\operatorname{sg}$ for stop-gradient.

\subsection{Token Objectives and Fixed Mixtures}
\label{sec:appendix-token-losses}

The per-position losses are target-token cross-entropy (CE), forward KL, and TV,
\begin{align}
  \ell_i^{\mathrm{CE}}&=-\log q_i(y_i),\\
  \ell_i^{\mathrm{KL}}&=\sum_{v\in\mathcal V}p_i(v)\log\frac{p_i(v)}{q_i(v)},\\
  \ell_i^{\mathrm{TV}}&=\frac12\sum_{v\in\mathcal V}|p_i(v)-q_i(v)|.
  \label{eq:appendix-token-losses}
\end{align}
For a fixed prefix, the expectation of CE over $y_i\sim p_i$ equals $H(p_i)+\ell_i^{\mathrm{KL}}$, although target-token CE and KL give different training signals on a finite sample. TV+KL uses $0.9\ell_i^{\mathrm{TV}}+0.1\ell_i^{\mathrm{KL}}$, and TV+CE uses $0.9\ell_i^{\mathrm{TV}}+0.1\ell_i^{\mathrm{CE}}$. TV+CE adapts DSpark's objective~\citep{DSpark}, which weights the L1 distance, $2\ell_i^{\mathrm{TV}}$, by 0.9.

TV, KL, CE, the fixed mixtures and LK all weight positions by DSpark's $h_i=\exp(-(i-1)/4)$ and normalize by the total weight,
\begin{equation}
  \mathcal R_h(\ell)=\frac{\sum_{b\in\mathcal B}\sum_{i=1}^{K}m_{bi}h_i\ell_{bi}}{\sum_{b\in\mathcal B}\sum_{i=1}^{K}m_{bi}h_i},
  \label{eq:appendix-weighted-reduction}
\end{equation}
where $\mathcal B$ indexes the training blocks in a batch and $m_{bi}\in\{0,1\}$ marks valid positions.

\subsection{LK and LK-log}

LK~\citep{LKLoss} mixes KL and TV with a weight set by the current acceptance. Let $\alpha_{bi}=1-\ell_{bi}^{\mathrm{TV}}$. As in LK, the weight is computed separately for each draft position from the mean acceptance at that position over the whole batch, and LK's $\eta=3$ is used,
\begin{equation}
  \bar\alpha_i=\frac{\sum_b m_{bi}\alpha_{bi}}{\sum_b m_{bi}},\qquad
  \lambda_i=\exp\bigl(-\eta\operatorname{sg}(\bar\alpha_i)\bigr),\qquad
  \ell_{bi}^{\mathrm{LK}}=\lambda_i\ell_{bi}^{\mathrm{KL}}+(1-\lambda_i)\ell_{bi}^{\mathrm{TV}},
  \label{eq:appendix-lk}
\end{equation}
with the mean taken across data-parallel workers and the loss reduced by \cref{eq:appendix-weighted-reduction}. LK-log is LK's negative log-acceptance loss, $\ell_{bi}^{\mathrm{LK\text{-}log}}=-\log\bigl(1-\ell_{bi}^{\mathrm{TV}}\bigr)$, with the argument of the logarithm floored at $10^{-6}$, and uses the same reduction.

\subsection{Confidence-Adaptive Token Weighting}

PARD-2's CAT~\citep{PARD2} weights each position by the target's confidence on the preceding training tokens,
\begin{equation}
  c_{bi}=\prod_{j<i}p_{bj}(y_{bj}),\qquad c_{b1}=1.
\end{equation}
We apply these weights to TV+KL in place of $h_i$, normalized by their total like the other weighted losses,
\begin{equation}
  \mathcal L_{\mathrm{CAT}}=\frac{\sum_{b,i}m_{bi}\operatorname{sg}(c_{bi})\bigl(0.9\ell_{bi}^{\mathrm{TV}}+0.1\ell_{bi}^{\mathrm{KL}}\bigr)}{\sum_{b,i}m_{bi}\operatorname{sg}(c_{bi})}.
  \label{eq:appendix-cat}
\end{equation}
PARD-2 itself applies the weights to CE plus distillation on its own drafter, so CAT here isolates the weighting.

\subsection{D-PACE and D-PAKL}

D-PACE~\citep{DPACE} weights CE by a surrogate of the expected acceptance length, $\sum_{m=1}^{K}\prod_{j\le m}q_j(y_j)$, built from the draft's confidence on the target token. With D-PACE's smoothing $a=0.5$ ($\alpha$ in their paper), let
\begin{equation}
  \widetilde q_i=(1-a)q_i(y_i)+a,\qquad
  \bar w_i=\sum_{m=i}^{K}\prod_{j=1}^{m}\widetilde q_j=\Bigl(\prod_{j\le i}\widetilde q_j\Bigr)\widetilde f_i,
  \label{eq:appendix-dpace-weights}
\end{equation}
where $\widetilde f_i=1+\sum_{m=i+1}^{K}\prod_{j=i+1}^{m}\widetilde q_j$ is the continuation value. The block loss is
\begin{equation}
  L_{\mathrm{D\text{-}PACE}}=\sum_{i=1}^{K}\operatorname{sg}(\bar w_i)\bigl(-\log q_i(y_i)\bigr).
\end{equation}
D-PAKL is D-PACE's KL variant~\citep{DPACE}. It replaces the confidence $q_i(y_i)$ by $\hat a_i=\exp(-C_i)$, where $C_i=-\sum_v p_i(v)\log q_i(v)$, and weights forward KL,
\begin{equation}
  L_{\mathrm{D\text{-}PAKL}}=\sum_{i=1}^{K}\operatorname{sg}\bigl(\bar w_i(\hat a)\bigr)\ell_i^{\mathrm{KL}},
\end{equation}
where $\bar w_i(\hat a)$ is \cref{eq:appendix-dpace-weights} with $\hat a_i$ in place of $q_i(y_i)$. Weighting $C_i$ instead of KL gives the same gradient, since the two differ by the target entropy. Both losses sum over positions and average over complete blocks, without normalizing by the weights.

\subsection{Window TV}

WTV~\citep{WTV} is a surrogate for the expected acceptance length of a window. It multiplies the acceptances $1-\TV(p_j,q_j)$ of teacher-forced positions, as if acceptance at each position did not depend on the earlier ones. With one window per training block and sampling temperature $T=1$, the block loss is
\begin{equation}
  L_{\mathrm{WTV}}=-\sum_{i=1}^{K}\prod_{j=1}^{i}\bigl(1-\operatorname{TV}(p_j,q_j)\bigr),
  \label{eq:appendix-wtv}
\end{equation}
averaged over complete blocks. Gradients pass through every TV term in every prefix product, so the multiplier on $\nabla_\theta\ell_i^{\mathrm{TV}}$ is
\begin{equation}
  \frac{\partial L_{\mathrm{WTV}}}{\partial\ell_i^{\mathrm{TV}}}
  =\sum_{r=i}^{K}\prod_{\substack{j\leq r\\j\ne i}}(1-\ell_j^{\mathrm{TV}}).
\end{equation}

\subsection{DBLast Acceptance-Length Loss}

DBLast-AL is the acceptance-length loss of \citet{DBLAST}, trained on our drafter without DBLast's latent-variable architecture. On a target-sampled block, let
\begin{equation}
  r_\ell=\prod_{i=1}^{\ell}\frac{q_i(y_i)}{p_i(y_i)},\qquad
  P_i=\prod_{j=1}^{i}\min\left\{1,\frac{p_j(y_j)}{q_j(y_j)}\right\},\qquad
  S_\ell=\sum_{i=1}^{\ell}P_i.
\end{equation}
Here $r_\ell$ is the prefix likelihood ratio, and $S_\ell$ is the expected acceptance length within the first $\ell$ positions. DBLast keeps the prefixes through the first one whose likelihood ratio falls below $\tau=0.1$,
\begin{equation}
  \ell_\tau=\min\bigl(\{\ell:r_\ell<\tau\}\cup\{K\}\bigr),
\end{equation}
and the block loss is
\begin{equation}
  L_{\mathrm{DBLast\text{-}AL}}=-\sum_{\ell=1}^{\ell_\tau}\bigl(\log r_\ell+\log S_\ell\bigr).
  \label{eq:appendix-dblast}
\end{equation}
The truncation index is chosen without gradient, and gradients pass through both log terms. Block losses are averaged over blocks with a valid prefix. As \citet{DBLAST} note, the logarithm and the truncation make this a surrogate, not an unbiased estimate of expected acceptance length.

\subsection{Block Verification Loss}
\label{sec:appendix-bv}

BV~\citep{BV} trains drafts for block verification~\citep{sun2025block}. On a target-sampled block, let
\begin{equation}
  r_i=\prod_{j=1}^{i}\frac{q_j(y_j)}{p_j(y_j)},\qquad
  \widetilde\alpha_i=\min\{1,r_1,\ldots,r_i\},\qquad
  r_0=\widetilde\alpha_0=1.
\end{equation}
Integrating over the current token gives BV's prefix score
\begin{equation}
  \bar\alpha_i=\sum_{v\in\mathcal V}\min\{\widetilde\alpha_{i-1}p_i(v),r_{i-1}q_i(v)\},
  \label{eq:appendix-bv-score}
\end{equation}
whose sum over $i$ is an unbiased estimate of the expected acceptance length under block verification. At the first position, $\bar\alpha_1=1-\operatorname{TV}(p_1,q_1)$. BV anneals how these scores are aggregated,
\begin{equation}
  g_i=\max\{\log\bar\alpha_i,-80\},\qquad
  w_i=\frac{\exp(\rho g_i)}{\sum_{j=1}^{K}\exp(\rho g_j)},\qquad
  L_{\mathrm{BV}}=-\sum_{i=1}^{K}\operatorname{sg}(w_i)g_i,
  \label{eq:appendix-bv-loss}
\end{equation}
with the weights held fixed in the backward pass and gradients passing through the prefix ratios and minima. We write $\rho$ for BV's $\beta$, to avoid a clash with TVP's power. BV raises it linearly from 0 to 1 over the first three of six epochs, and we likewise raise $\rho$ over the first half of training and then hold it at 1, where the gradient is that of the blockwise log loss $-\log(K^{-1}\sum_i\bar\alpha_i)$. BV floors the log scores at $-80$ once $\rho=1$, and we apply the same floor throughout. Block losses are averaged over complete blocks. We evaluate BV-trained drafts both with the tokenwise verifier used for the other losses and with the block verifier BV is designed for (\cref{tab:drafts}).

\subsection{EAL Loss}
\label{sec:appendix-eal-loss}

The EAL loss optimizes EAL directly. For a fixed anchor, let $Y_{1:K}$ be sampled from the target, with $p_i$ and $q_i$ evaluated on the same sampled prefix, and let $B_{i-1}=\prod_{j<i}\min\{1,q_j(Y_j)/p_j(Y_j)\}$ as in \cref{sec:expected-acceptance-length}. The loss is
\begin{equation}
  \mathcal L_{\mathrm{EAL}}=\mathbb E_{Y\sim p}\left[\frac1K\sum_{i=1}^{K}(K-i+1)B_{i-1}\operatorname{TV}(p_i,q_i)\right],
  \label{eq:appendix-eal-loss}
\end{equation}
which equals the EAL deficit $1-\EAL/K$ by \cref{eq:tvp-exact-tv-appendix}. The ratio is $q/p$ rather than the verifier's $p/q$ because the prefixes are sampled from the target. Gradients pass through both $B_{i-1}$ and TV, and block losses are averaged over complete blocks.

\subsection{DFlash2 Selector Objectives}
\label{sec:appendix-selector-losses}

DFlash2 drafts each position from a restricted candidate set~\citep{inco2026dflash2}. At one position, let $C$ be the native top-$M$ candidates, $s$ the selector's softmax over $C$, which is zero outside $C$, and $m=\sum_{v\in C}p(v)$ the target mass inside $C$. The selector's acceptance and TV are
\begin{equation}
  A_C=\sum_{v\in C}\min\{p(v),s(v)\},\qquad
  \operatorname{TV}_C=1-A_C=(1-m)+\sum_{v\in C}[p(v)-s(v)]_+.
\end{equation}
We do not renormalize $p$ on $C$, so target mass outside $C$ always counts as rejected. Forward KL to $s$ is infinite whenever $m<1$, so the KL term uses the target restricted to $C$, weighted by its mass,
\begin{equation}
  \operatorname{KL}_C=m\operatorname{KL}\bigl(p(\cdot\mid C)\|s\bigr),
\end{equation}
which is zero when $m=0$. TV+KL on the selector is $0.9\operatorname{TV}_C+0.1\operatorname{KL}_C$, and LK uses the same terms with $\lambda_i=\exp(-\eta\operatorname{sg}(\bar A_{C,i}))$, where $\bar A_{C,i}$ is the mean acceptance at draft position $i$.

For TVP, we apply $D_\beta$ inside $C$ and add the missing mass,
\begin{equation}
  D_{\beta,C}(p,s)=(1-m)+
  \begin{cases}
    \displaystyle\frac1\beta\sum_{v\in C}p(v)\left[1-\min\left\{1,\frac{s(v)}{p(v)}\right\}^{\!\beta}\right], & 0<\beta\le1,\\[6pt]
    \displaystyle\sum_{v\in C}p(v)\left[\log\frac{p(v)}{s(v)}\right]_+, & \beta=0,
  \end{cases}
  \label{eq:appendix-selector-tvp}
\end{equation}
so that $D_{1,C}=\operatorname{TV}_C$ and the $\beta=0$ case stays finite. The target probabilities are detached, terms with $p(v)=0$ contribute zero, and $1-m$ has no gradient for a fixed candidate set. We compute the power term as $-\operatorname{expm1}(\beta\min\{0,\log s-\log p\})/\beta$, which stays accurate for small $\beta$.

For CE, the target token may fall outside $C$. One recipe inserts it into the candidate set, and the other keeps the native set and drops CE when the token is absent, without renormalizing over the remaining positions. In these comparisons, the backbone keeps its own TV+KL loss. The full-vocabulary variants instead combine the backbone and selector scores into a distribution over the whole vocabulary during training and apply the loss to that distribution. Inference always uses the native candidate set.

%% file: sections/appendix/appendix_experiments.tex
\section{Experimental and Implementation Details}
\label{sec:appendix-implementation}

\paragraph{Data.} We use the Open-PerfectBlend dataset for training. In each epoch, the target generates one thinking-on completion per prompt at $T=1$ with up to 4{,}096 new tokens, using a new seed, so no completion is repeated. We stream the target's hidden states to the trainer live using Baseten's capture pipeline~\citep{baseten2026livedraft}. The data is sent in the same order, and the generations are identical for each trainer, ensuring consistency.

\paragraph{Optimization.} All runs use AdamW with no weight decay, a 4\% linear warmup followed by cosine decay to zero, gradient clipping at 1.0, and bfloat16. Each sequence contributes 512 anchors. Per-token objectives share DSpark's position weights $\exp(-(i-1)/4)$, and DSpark's confidence head keeps its own loss in every run. TVP anneals $\beta$ linearly from 0 to 1 over the first 80\% of training and holds it at 1 afterwards. \Cref{tab:hparams} lists the remaining settings.

\paragraph{Evaluation.} DSpark heads are served in SGLang~\citep{zheng2024sglang} with a block of 8 tokens (7 draft tokens) and rejection-sampling verification. The suite has 4{,}240 prompts across nine benchmarks, with up to 8{,}192 new tokens, and $\tau$ is the unweighted mean over benchmarks, pooling two samples per prompt at $T=1$.

\begin{table}[h]
  \caption{Training settings. FT starts from the released head, and PT from random initialization.}
  \label{tab:hparams}
  \centering
  \small
  \begin{tabular}{@{}llrrr@{}}
    \toprule
    Setting & Init & Learning rate & Sequences per step & Epochs \\
    \midrule
    Qwen3-8B DSpark, FT & released & $2\times10^{-4}$ & 16 & 1 \\
    Qwen3-8B DSpark, FT, $B=15$ & released & $2\times10^{-4}$ & 16 & 1 \\
    Qwen3-8B DSpark, FT (\cref{tab:budget,tab:beta-start,tab:causal,tab:curves}) & released & $10^{-4}$ & 16 & 1 to 5 \\
    Qwen3-14B DSpark, FT & released & $2\times10^{-4}$ & 16 & 1 \\
    Qwen3-8B DSpark, PT & random & $6\times10^{-4}$ & 32 & 2 to 5 \\
    Qwen3.8-27B DFlash2, PT & random & $6\times10^{-4}$ & 32 & 2 \\
    Qwen3.8-27B DFlash2, FT & released & $5\times10^{-5}$ & 16 & 2 \\
    Qwen3-8B Eagle3, FT & released & $10^{-4}$ & 15 & $<1$ \\
    \bottomrule
  \end{tabular}
\end{table}

%% file: sections/appendix/appendix_results.tex
\section{Additional Results and Ablations}
\label{sec:appendix-results}

\subsection{Evaluation and seed noise}
\label{sec:appendix-noise}
Every $\tau$ in the paper averages 4{,}240 prompts with thinking enabled, generations of up to 8{,}192 new tokens, and two samples per prompt at $T=1$. Each $\Delta_{\mathrm{TV}}$ carries a 95\% paired bootstrap confidence interval (CI) over prompts, typically $\pm0.2$ to $0.3$ points.

Three seeds of the one-epoch TVP fine-tuning run give $\tau=4.478$, 4.478 and 4.490, a standard deviation of 0.007 (0.15\%), and two seeds at $B=15$ give 4.991 and 4.984. Runs of TVP, TV+KL, CAT and fixed $\beta=0.5$ repeated on a second data draw differ by 0.01 to 0.14\%. In pretraining we similarly see a difference of 0.57\%. We therefore read fine-tuning differences below about 0.3\% and pretraining differences below 0.6\% as ties.

\subsection{Learning-rate sweep}
\label{sec:appendix-lr}
\Cref{tab:lr} trains six objectives at five learning rates. TVP has the highest $\tau$ at every learning rate, and KL the lowest. The best learning rate is $2\times10^{-4}$ or $4\times10^{-4}$ for every objective, and $2\times10^{-4}$ is within 0.2\% of the best for each, so the main tables use it throughout. The order of TV, TV+KL and LK, which stay within 1.1\% of each other, changes with the learning rate. TV is second at $2\times10^{-4}$ and below, and fourth at higher learning rates. The TVP entries are single runs, so the $2\times10^{-4}$ value (4.469) differs slightly from the three-seed mean in \cref{tab:ft8b} (4.482). The ablations in \cref{tab:budget,tab:beta-start,tab:causal,tab:curves} use $10^{-4}$. Going from $10^{-4}$ to $2\times10^{-4}$ raises every objective by 0.4 to 1.0\% and keeps TVP first, and TVP's margin over TV grows from $+0.9\%$ to $+1.2\%$, so these ablations, if anything, understate TVP.
\input{sections/appendix/results_tables/tab_lr}

\subsection{TVP schedule ablations}
\label{sec:appendix-beta}
\input{sections/tables/tab_beta}

\subsection{Greedy decoding}
\label{sec:appendix-greedy}
Under greedy decoding, a draft token is accepted when it matches the target's argmax, so $1-\TV(p,q)$ no longer gives the acceptance probability. Greedy acceptance can instead be bounded through KL~\citep{GreedyTheory}, while TV measures the overlap of the whole distribution, which sets acceptance under sampling but not under greedy decoding. This is consistent with objectives with a KL or CE term doing relatively better at $T=0$ (\cref{tab:greedy}). In fine-tuning, TV (4.780) falls to the level of KL (4.773), while LK and mixtures with a KL or CE term (TV+KL, TV+CE) rise above it. In pretraining, KL itself beats TV (4.964 against 4.864) and LK is best (5.088). TVP stays near the top, tied with TV+KL in fine-tuning (4.857) and 1.0\% below LK in pretraining (5.039), consistent with its early KL-like phase. At one request every trained head reaches $3.21$ to $3.31\times$, within each other's intervals, against $2.63\times$ for the released head.
\input{sections/appendix/results_tables/tab_greedy}

\subsection{Optimizing EAL directly}
\label{sec:appendix-causal}
The EAL loss (\cref{eq:appendix-eal-loss}) weights each position's TV by the survival of its target-sampled prefix, so its expected value is exactly the EAL deficit $1-\EAL/K$ (\cref{eq:tvp-exact-tv-appendix}). It is the natural objective for optimizing EAL itself, yet it trains worse than TV (\cref{tab:causal}). When fine-tuning, the EAL loss is 1.9\% below TV. From scratch, where early positions are rarely accepted, it falls 9.5\% below TV after two epochs. Its loss allocation shows why. In fine-tuning, the EAL loss puts 32\% of its loss on the first position, which already accepts about 0.85, and 2\% on the seventh, while the fixed-weight objectives put 13 to 17\% on every position. This is the prefix attenuation of \cref{sec:prefix-attenuation}.
\input{sections/appendix/results_tables/tab_causal}

\subsection{Acceptance by target entropy}
\label{sec:appendix-entropy}
\Cref{tab:entropy-full} extends \cref{tab:entropy} to all seven target-entropy buckets and adds Qwen3-8B, which shows the same pattern. On both targets, TVP's acceptance gain over TV grows steadily with target entropy, from zero below 0.05 nats to about $+0.018$ above 3 nats. KL's acceptance rate is below TV's in every bucket under 1 nat, which together hold 86 to 87\% of positions. Its lead over TVP in the 1 to 2 nat bucket is within noise, so KL only leads above 2 nats, on about 1\% of positions. KL also has the lowest forward KL in every bucket, so a lower forward KL does not give a higher acceptance rate.
\input{sections/appendix/results_tables/tab_entropy_full}

\subsection{Training curves}
\label{sec:appendix-curves}
\Cref{tab:curves} lists every objective at every evaluated checkpoint of the two 5-epoch runs in \cref{fig:curves}.
\input{sections/appendix/results_tables/tab_curves}

\subsection{DFlash2 fine-tuning}
\label{sec:appendix-dflash2-ft}
\Cref{tab:dflash2-ft} fine-tunes the released Qwen3.8-27B DFlash2 head. We use a learning rate of $5\times10^{-5}$, because at $10^{-4}$ every objective first moved away from the released head. Every TV-family objective improves the released head, by 2.3 to 2.9\%, while SpecForge's CE recipe ties it and TorchSpec's degrades it by 5.0\%. TVP does not improve on TV here. With the same support it is $+0.3\%$ on the full vocabulary, within noise, and $-0.6\%$ on the top-16 set, which makes this the one setting where TV has the highest $\tau$.
\input{sections/appendix/results_tables/tab_dflash2_ft}

%% file: sections/appendix/results_tables/tab_lr.tex
\begin{table}[h]
  \caption{Learning-rate sweep, Qwen3-8B DSpark fine-tuning for one epoch, $\tau$ at $T=1$.}
  \label{tab:lr}
  \centering
  \small
  \begin{tabular}{@{}lrrrrr@{}}
    \toprule
    Objective & $5\times10^{-5}$ & $10^{-4}$ & $2\times10^{-4}$ & $4\times10^{-4}$ & $6\times10^{-4}$ \\
    \midrule
    TV    & 4.344 & 4.399 & 4.415 & 4.384 & 4.367 \\
    KL    & 4.036 & 4.100 & 4.143 & 4.140 & 4.121 \\
    TV+KL & 4.313 & 4.370 & 4.404 & 4.413 & 4.400 \\
    \midrule
    LK     & 4.297 & 4.366 & 4.401 & 4.404 & 4.392 \\
    D-PACE & 4.200 & 4.244 & 4.266 & 4.268 & 4.240 \\
    \midrule
    TVP & \textbf{4.379} & \textbf{4.437} & \textbf{4.469} & \textbf{4.470} & \textbf{4.456} \\
    \bottomrule
  \end{tabular}
\end{table}

%% file: sections/tables/tab_beta.tex
\begin{table}[t]
  \caption{TVP's $\beta$ schedule, $\Delta$ is $\tau$ at $T=1$ relative to the $\beta$ 0$\to$1 anneal (all 95\% CIs are $\pm0.2$). (a) Starting value and fixed $\beta$. FT is Qwen3-8B fine-tuned for one epoch at learning rate $10^{-4}$. PT is Qwen3-8B pretrained for two epochs. (b) Shape of the 0$\to$1 anneal, Qwen3-8B fine-tuned for one epoch at learning rate $2\times10^{-4}$.}
  \label{tab:beta}
  \centering
  \footnotesize
  \setlength{\tabcolsep}{4pt}
  \begin{subtable}[t]{0.47\columnwidth}
    \centering
    \caption{Start and fixed $\beta$, $\Delta$ (\%).}
    \label{tab:beta-start}
    \begin{tabular}{@{}lrr@{}}
      \toprule
      Schedule & FT & PT \\
      \midrule
      $\beta$ 0$\to$1    & \tvref & \tvref \\
      $\beta$ 0.25$\to$1 & $+0.0$ & $-0.8$ \\
      $\beta$ 0.5$\to$1  & $-0.3$ & $-1.8$ \\
      $\beta$ 0.75$\to$1 & $-0.6$ & $-3.7$ \\
      \midrule
      $\beta=0.5$        & $-1.4$ & $-2.0$ \\
      $\beta=1$ (TV)     & $-0.8$ & $-5.1$ \\
      \bottomrule
    \end{tabular}
  \end{subtable}\hfill
  \begin{subtable}[t]{0.51\columnwidth}
    \centering
    \caption{Shape of the anneal.}
    \label{tab:beta-shape}
    \begin{tabular}{@{}lrr@{}}
      \toprule
      Schedule & $\tau$ & $\Delta$ (\%) \\
      \midrule
      Linear, by 80\%  & \textbf{4.482} & \tvref \\
      Linear, by 100\% & \textbf{4.482} & $-0.0$ \\
      Linear, by 40\%  & 4.457 & $-0.6$ \\
      \midrule
      Step at 40\%     & 4.462 & $-0.5$ \\
      Step at 80\%     & 4.425 & $-1.3$ \\
      \midrule
      $\beta=0$        & 4.120 & $-8.1$ \\
      $\beta$ 1$\to$0  & 4.109 & $-8.3$ \\
      \bottomrule
    \end{tabular}
  \end{subtable}
\end{table}

%% file: sections/appendix/results_tables/tab_greedy.tex
\begin{table}[h]
  \caption{Greedy decoding ($T=0$) for the Qwen3-8B DSpark runs of \cref{tab:ft8b,tab:pt8b}. Speedup is over Qwen3-8B alone at one request, with 95\% CIs.}
  \label{tab:greedy}
  \centering
  \small
  \begin{tabular}[t]{@{}lrr@{}}
    \multicolumn{3}{@{}l}{(a) Fine-tuning, one epoch} \\
    \toprule
    Objective & $\tau$ & Speedup \\
    \midrule
    Released head & 3.894 & \spd{2.63}{0.05} \\
    \midrule
    TV            & 4.780 & \spd{3.23}{0.10} \\
    KL            & 4.773 & \spd{3.25}{0.10} \\
    CE            & 4.711 & \spd{3.21}{0.10} \\
    TV+KL         & \textbf{4.857} & \spd{3.27}{0.10} \\
    TV+CE         & 4.835 & \textbf{\spd{3.31}{0.10}} \\
    \midrule
    LK            & 4.836 & \spd{3.28}{0.09} \\
    LK-log        & 4.780 & \spd{3.25}{0.10} \\
    D-PACE        & 4.805 & \spd{3.26}{0.09} \\
    WTV           & 4.809 & \spd{3.23}{0.10} \\
    BV            & 4.836 & \spd{3.26}{0.10} \\
    \midrule
    TVP, $\beta=0$ & 4.814 & \spd{3.27}{0.09} \\
    TVP            & \textbf{4.857} & \spd{3.30}{0.10} \\
    \bottomrule
  \end{tabular}\hspace{2.5em}
  \begin{tabular}[t]{@{}lr@{}}
    \multicolumn{2}{@{}l}{(b) Pretraining, five epochs} \\
    \toprule
    Objective & $\tau$ \\
    \midrule
    TV    & 4.864 \\
    KL    & 4.964 \\
    TV+KL & 5.036 \\
    \midrule
    CAT       & 5.031 \\
    LK        & \textbf{5.088} \\
    D-PACE    & 4.946 \\
    D-PAKL    & 5.014 \\
    DBLast-AL & 4.944 \\
    WTV       & 4.763 \\
    \midrule
    TVP, $\beta=0.5$   & 5.002 \\
    TVP, \anneal{0.5}  & 5.011 \\
    TVP, \anneal{0.25} & 5.022 \\
    TVP                & 5.039 \\
    \bottomrule
  \end{tabular}
\end{table}

%% file: sections/appendix/results_tables/tab_causal.tex
\begin{table}[h]
  \caption{Optimizing EAL directly. The EAL loss equals the EAL deficit $1-\EAL/K$ (\cref{eq:appendix-eal-loss}). Qwen3-8B DSpark, $\tau$ at $T=1$, with the change relative to TV in parentheses. Fine-tuning runs one epoch at learning rate $10^{-4}$ and pretraining two epochs at $6\times10^{-4}$.}
  \label{tab:causal}
  \centering
  \small
  \begin{tabular}{@{}lrrr@{}}
    \toprule
    Regime & EAL loss & TV & TVP \\
    \midrule
    Fine-tuning & 4.314 ($-1.9\%$) & 4.399 & \textbf{4.437} ($+0.9\%$) \\
    Pretraining & 3.815 ($-9.5\%$) & 4.217 & \textbf{4.417} ($+4.7\%$) \\
    \bottomrule
  \end{tabular}
\end{table}

%% file: sections/appendix/results_tables/tab_entropy_full.tex
\begin{table}[h]
  \caption{Acceptance rate, draft entropy and forward KL for every target-entropy bucket (nats), Qwen3-8B and Qwen3-14B DSpark fine-tuning, last 5\% of training. Acceptance is the mean of $1-\TV(p_i,q_i)$ over draft positions. Bold marks the best acceptance in each row, and underlined entries are within noise (0.01) of it.}
  \label{tab:entropy-full}
  \centering
  \small
  \begin{tabular}{@{}lrccccccccc@{}}
    \toprule
    & & \multicolumn{3}{c}{Acceptance} & \multicolumn{3}{c}{Draft entropy} & \multicolumn{3}{c}{Forward KL} \\
    \cmidrule(lr){3-5} \cmidrule(lr){6-8} \cmidrule(l){9-11}
    Target entropy & Share & TVP & TV & KL & TVP & TV & KL & TVP & TV & KL \\
    \midrule
    \multicolumn{11}{@{}l}{\emph{Qwen3-8B}} \\
    $<0.05$    & 47.3\% & \underline{0.882} & \textbf{0.883} & 0.826 & 0.18 & 0.15 & 0.62 & 0.63 & 0.75 & 0.44 \\
    0.05--0.2  & 9.5\%  & \textbf{0.710} & \underline{0.709} & 0.619 & 0.50 & 0.43 & 1.36 & 1.41 & 1.67 & 1.00 \\
    0.2--0.5   & 11.3\% & \textbf{0.647} & \underline{0.642} & 0.581 & 0.64 & 0.57 & 1.54 & 1.65 & 1.96 & 1.10 \\
    0.5--1     & 18.4\% & \textbf{0.580} & 0.570 & 0.559 & 0.84 & 0.76 & 1.73 & 1.88 & 2.26 & 1.16 \\
    1--2       & 12.6\% & \underline{0.485} & 0.472 & \textbf{0.492} & 1.24 & 1.15 & 2.28 & 2.38 & 2.83 & 1.44 \\
    2--3       & 0.9\%  & 0.344 & 0.327 & \textbf{0.379} & 1.82 & 1.69 & 3.27 & 3.45 & 4.01 & 2.05 \\
    $>3$       & 0.1\%  & 0.266 & 0.249 & \textbf{0.322} & 2.60 & 2.43 & 4.50 & 4.05 & 4.62 & 2.38 \\
    \midrule
    \multicolumn{11}{@{}l}{\emph{Qwen3-14B}} \\
    $<0.05$    & 47.6\% & \underline{0.876} & \textbf{0.877} & 0.819 & 0.19 & 0.16 & 0.64 & 0.66 & 0.79 & 0.46 \\
    0.05--0.2  & 9.1\%  & \textbf{0.696} & \underline{0.695} & 0.605 & 0.52 & 0.45 & 1.42 & 1.50 & 1.78 & 1.07 \\
    0.2--0.5   & 10.9\% & \textbf{0.636} & \underline{0.631} & 0.569 & 0.67 & 0.59 & 1.60 & 1.73 & 2.06 & 1.16 \\
    0.5--1     & 18.2\% & \textbf{0.571} & 0.561 & 0.548 & 0.85 & 0.77 & 1.79 & 1.95 & 2.35 & 1.22 \\
    1--2       & 13.0\% & \underline{0.474} & 0.460 & \textbf{0.479} & 1.27 & 1.17 & 2.36 & 2.48 & 2.95 & 1.51 \\
    2--3       & 1.1\%  & 0.344 & 0.327 & \textbf{0.377} & 1.84 & 1.70 & 3.30 & 3.45 & 4.02 & 2.07 \\
    $>3$       & 0.1\%  & 0.257 & 0.239 & \textbf{0.312} & 2.58 & 2.39 & 4.51 & 4.14 & 4.75 & 2.47 \\
    \bottomrule
  \end{tabular}
\end{table}

%% file: sections/appendix/results_tables/tab_curves.tex
\begin{table}[h]
  \caption{$\tau$ at $T=1$ relative to TV (\%) at every evaluated checkpoint of the 5-epoch Qwen3-8B DSpark runs. The second row gives TV's own $\tau$.}
  \label{tab:curves}
  \centering
  \small
  \begin{tabular}{@{}lrrrrrrr@{}}
    \toprule
    \multicolumn{8}{@{}l}{\textit{Fine-tuning, 5 epochs, learning rate $10^{-4}$}} \\
    \midrule
    Training (\%) & 38 & 54 & 63 & 72 & 81 & 90 & 100 \\
    TV, $\tau$ & 4.441 & 4.523 & 4.569 & 4.598 & 4.609 & 4.626 & 4.633 \\
    \midrule
    KL & $-6.3$ & $-6.3$ & $-6.3$ & $-6.2$ & $-6.2$ & $-6.0$ & $-6.3$ \\
    TV+KL & $+0.0$ & $-0.1$ & $-0.3$ & $-0.4$ & $-0.2$ & $-0.3$ & $-0.3$ \\
    CAT & $-0.3$ & $-0.4$ & $-0.5$ & $-0.4$ & $-0.1$ & $-0.4$ & $-0.4$ \\
    LK & $-0.2$ & $-0.5$ & $-0.4$ & $-0.5$ & $-0.3$ & $-0.4$ & $-0.4$ \\
    D-PACE & $-3.6$ & $-3.6$ & $-3.4$ & $-3.3$ & $-3.1$ & $-3.4$ & $-3.3$ \\
    D-PAKL & $-2.8$ & $-2.9$ & $-3.0$ & $-3.1$ & $-2.8$ & $-2.9$ & $-2.9$ \\
    TVP, $\beta=0.5$ & $-0.1$ & $-0.2$ & $-0.2$ & $-0.3$ & $-0.0$ & $-0.1$ & $-0.4$ \\
    TVP, \anneal{0.5} & $+0.6$ & $+0.6$ & $+0.8$ & $+0.8$ & $+0.9$ & $+0.9$ & $+0.7$ \\
    TVP & $-0.3$ & $+0.6$ & $+0.8$ & $+1.1$ & $+1.4$ & $+1.1$ & $+1.1$ \\
    \bottomrule
  \end{tabular}
  \par\vspace{10pt}
  \begin{tabular}{@{}lrrrrrr@{}}
    \toprule
    \multicolumn{7}{@{}l}{\textit{Pretraining from scratch, 5 epochs}} \\
    \midrule
    Training (\%) & 18 & 36 & 54 & 72 & 90 & 100 \\
    TV, $\tau$ & 3.880 & 4.149 & 4.311 & 4.419 & 4.475 & 4.465 \\
    \midrule
    KL & $-3.8$ & $-3.7$ & $-3.6$ & $-3.4$ & $-3.3$ & $-2.9$ \\
    TV+KL & $+3.7$ & $+3.2$ & $+2.9$ & $+2.6$ & $+2.6$ & $+2.8$ \\
    CAT & $+2.5$ & $+2.1$ & $+2.1$ & $+2.0$ & $+1.9$ & $+2.2$ \\
    LK & $+2.9$ & $+3.1$ & $+2.9$ & $+2.8$ & $+2.5$ & $+3.0$ \\
    D-PACE & $-0.1$ & $-1.1$ & $-1.1$ & $-1.0$ & $-0.9$ & $-0.7$ \\
    D-PAKL & $+1.3$ & $+0.3$ & $+0.4$ & $-0.2$ & $-0.1$ & $+0.3$ \\
    DBLast-AL & $-0.5$ & $-1.1$ & $-1.3$ & $-1.1$ & $-1.2$ & $-0.9$ \\
    WTV & $-3.6$ & $-3.0$ & $-2.3$ & $-2.3$ & $-2.1$ & $-2.0$ \\
    TVP, $\beta=0.5$ & $+2.7$ & $+2.4$ & $+2.3$ & $+1.9$ & $+2.0$ & $+2.3$ \\
    TVP, \anneal{0.5} & $+3.7$ & $+3.2$ & $+2.9$ & $+2.8$ & $+2.5$ & $+2.9$ \\
    TVP, \anneal{0.25} & $+3.2$ & $+3.7$ & $+3.8$ & $+3.7$ & $+3.5$ & $+3.8$ \\
    TVP & $+0.7$ & $+3.2$ & $+3.9$ & $+4.1$ & $+3.8$ & $+4.2$ \\
    \bottomrule
  \end{tabular}
\end{table}

%% file: sections/appendix/results_tables/tab_dflash2_ft.tex
\begin{table}[h]
  \caption{Qwen3.8-27B DFlash2 fine-tuned from the released head for two epochs at learning rate $5\times10^{-5}$, $\tau$ at $T=1$ and its change relative to the released head. Top-16 variants restrict the selector objective to the native candidate set (\cref{sec:appendix-selector-losses}). With the same support, TVP is $+0.3\%$ over TV on the full vocabulary (not significant) and $-0.6\%$ on the top-16 set. DFlash2 evaluations are noisier than DSpark's, with 95\% CIs of about $\pm0.4$ to $0.5$ points.}
  \label{tab:dflash2-ft}
  \centering
  \small
  \begin{tabular}{@{}lrr@{}}
    \toprule
    Objective & $\tau$ & vs.\ released (\%) \\
    \midrule
    Released head & 3.741 & \\
    \midrule
    TorchSpec CE  & 3.553 & $-5.0$ \\
    SpecForge CE  & 3.736 & $-0.1$ \\
    TV            & 3.828 & $+2.3$ \\
    TV, top-16    & \textbf{3.849} & $+2.9$ \\
    \midrule
    TVP, top-16   & 3.827 & $+2.3$ \\
    TVP           & 3.839 & $+2.6$ \\
    \bottomrule
  \end{tabular}
\end{table}